\documentclass[twoside,11pt]{article}

\usepackage{jmlr2e}
\usepackage{times}
\usepackage{epsfig}
\usepackage{graphicx}
\usepackage{amsmath}
\usepackage{amssymb}
\usepackage[usenames,dvipsnames,svgnames,table]{xcolor}
\graphicspath{{./}{./images/}}

\usepackage[ruled,vlined]{algorithm2e}
\usepackage{booktabs,balance,adjustbox,rotating,array,enumitem,mathtools}
\usepackage{xspace,url,subcaption,multirow}
\usepackage[backref,colorlinks=true,linkcolor=blue,citecolor=blue,urlcolor=blue]{hyperref}

\makeatletter
\DeclareRobustCommand\onedot{\futurelet\@let@token\@onedot}
\def\@onedot{\ifx\@let@token.\else.\null\fi\xspace}

\def\eg{\emph{e.g}\onedot} 
\def\ie{\emph{i.e}\onedot}

\def\etal{\emph{et al}\onedot}
\makeatother

\def\Vec#1{{\boldsymbol{#1}}}
\def\Mat#1{{\boldsymbol{#1}}}

\def\SPD#1{\mathcal{S}_{++}^{#1}}

\def\ST#1#2{\mathrm{St}({#2},{#1})}

\newcommand{\tr}{\mathop{\rm  Tr}\nolimits}
\newcommand{\expm}{\mathop{\rm  expm}\nolimits}

\newcommand{\subject}{\mathrm{s.t.}}

\begin{document}

\title{Learning an Invariant Hilbert Space for Domain Adaptation}

\author{\name Samitha Herath \email samitha.herath@data61.csiro.au \\
       \addr College of Engineering and Computer Science\\
      Australian National University \& Data61, CSIRO\\
      Canberra, Australia
       \AND
       \name Mehrtash Harandi \email mehrtash.harandi@data61.csiro.au \\
       \addr College of Engineering and Computer Science\\
      Australian National University \& Data61, CSIRO\\
      Canberra, Australia
        \AND
       \name Fatih Porikli \email fatih.porikli@anu.edu.au \\
       \addr College of Engineering and Computer Science\\
      Australian National University\\
      Canberra, Australia}

\maketitle

\begin{abstract}
This paper introduces a learning scheme to construct a Hilbert space (i.e., a vector space along its inner product) to address both unsupervised and semi-supervised domain adaptation problems. This is achieved by learning projections from each domain to a latent 
space along the Mahalanobis metric of the latent space to simultaneously minimizing a notion of domain variance while maximizing a measure of discriminatory power. In particular, we make use of the Riemannian optimization techniques to match statistical properties (e.g., first and second order statistics) between samples projected into the latent space from different domains. 
Upon availability of class labels, we further deem samples sharing the same label to form more compact clusters while pulling away samples coming from different classes.We extensively evaluate and contrast our proposal against state-of-the-art methods  
for the task of visual domain adaptation using both handcrafted and deep-net features. Our experiments show that even with a simple nearest neighbor classifier, the proposed method can outperform several state-of-the-art methods benefitting from more involved classification schemes.
\end{abstract}

\section{Introduction}

This paper presents a learning algorithm to address  both 
\emph{unsupervised}~\citep{Gong_CVPR12,Basura_ICCV13,SUN_Return15} and 
\emph{semi-supervised}~\citep{Hoffman_IJCV14,Duan_ICML12,Tsai_CVPR16} domain adaptation problems. 
Our goal here is to learn a latent space in which domain disparities are minimized. We show such a space can be learned by first matching the statistical properties of the projected domains (\eg, covariance matrices), and then adapting the Mahalanobis metric of the latent space to the labeled data, \ie, minimizing the distances between 
pairs sharing the same class label while pulling away samples with different class labels.
We develop a geometrical solution to jointly learn projections 
onto the latent space and the Mahalanobis metric there by making use of the concepts of Riemannian geometry.

Thanks to deep learning, we are witnessing rapid growth in classification accuracy  of the imaging techniques if substantial amount of 
labeled data is provided~\citep{Krizhevsky_NIPS12,Simonyan_VGG14,He_CVPR16,Herath_Survey17}. However, harnessing the attained knowledge into a new application with limited labeled data (or even without having labels) is far beyond clear~\citep{Koniusz_CVPR16,Long_Deep16,Ganin_ICML15,Tzeng_Deep14,Qiang_CVPR15}.
To make things even more complicated, due to the inherit \emph{bias} of datasets~\citep{Torralba_CVPR11,Shimodaira_JSPI00}, straightforward use of large amount of auxiliary data does not necessarily assure improved performances. For example, the ImageNet~\citep{ILSVRC15} data is hardly useful for an application designed to classify images captured by a mobile phone camera. 
Domain adaptation is the science of reducing such undesired effects in transferring knowledge from the available auxiliary resources 
to a new problem.

The most natural solution to the problem of DA  is by identifying the structure of a common space that minimizes a notion of 
domain mismatch. Once such a space is obtained, one can  design a classifier in it, 
hoping that the classifier will perform equally well across the domains as the domain mismatched is minimized. 
Towards this end, several studies assume  that either \textbf{1.} a subspace of the target\footnote{In DA terminology target domain refers to the data directly related to the task. Source domain data is used as the auxiliary data for knowledge transferring.} domain is the right space to perform DA and learn how the source domain should be mapped onto it~\citep{Saenko_ECCV10,Tsai_CVPR16},
or \textbf{2.} subspaces obtained from both source and target domains are equally important for classification, hence trying
to either learn their evolution~\citep{Gopalan_ICCV11,Gong_CVPR12}  or similarity measure~\citep{Shi_ICDM10,Wang_IJCAI11,Duan_ICML12}.

Objectively speaking, a common practice in many solutions including the aforementioned methods, 
is to simplify the learning problem by separating the two elements of it. That is,
the algorithm starts by fixing a space (\eg, source subspace of~\cite{Basura_ICCV13,Tsai_CVPR16}),
and learns how to transfer the knowledge from domains accordingly. A curious mind may ask why should we resort to a 
predefined and fixed space in the first place.

In this paper, we propose a learning scheme that avoids such a separation. That is, we do not assume that 
a space or a transformation, apriori is known and fixed for DA. In essence, we propose to 
learn the structure of a Hilbert space (\ie, its metric) along the transformations required to map the domains onto it jointly.

\begin{figure*}[!t]
\centering
\includegraphics[width=0.9\textwidth]{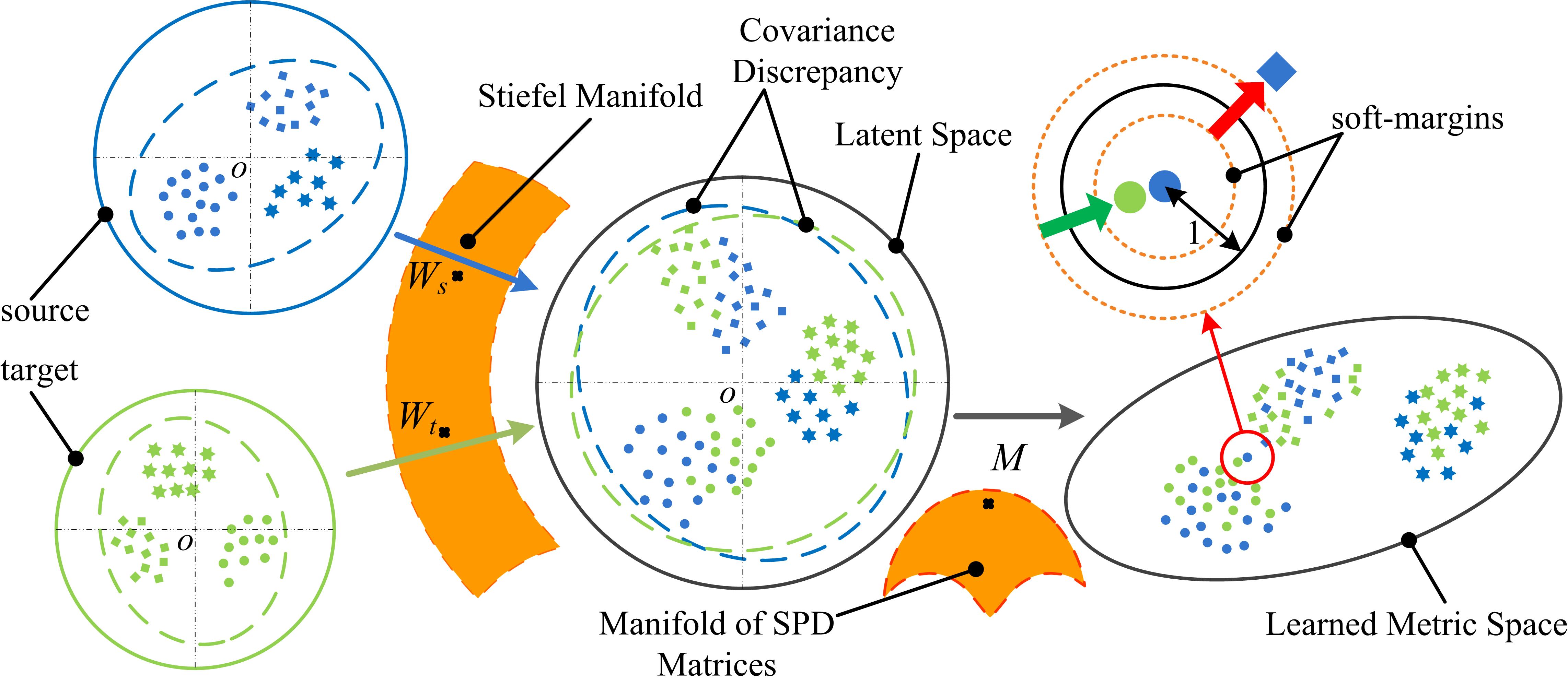}
\caption{\textbf{A conceptual diagram of our proposal.} 
The marker shape represents the instance labels and color represents their original domains. Both source and target domains are mapped to a latent space using the transformations $\Mat{W}_s$ and $\Mat{W}_t$. The metric, $\Mat{M}$ defined in the latent space is learned to maximize the discriminative power of samples in it. Indicated by dashed ellipsoids are the domain distributions. The statistical loss of our cost function aims to reduce such discrepancies within the latent space. Our learning scheme identifies the transformations $\Mat{W}_s$ and $\Mat{W}_t$ and the metric $\Mat{M}$ jointly. This figure is best viewed in color.}
\label{fig:mainDiag}
\end{figure*}

This is achieved through the following contributions,
\begin{itemize}
\item [(i)] We propose to learn the structure of a latent space, along its associated mappings from the source and target domains 
to address both problems of unsupervised and semi-supervised DA. 

\item [(ii)] Towards this end, we propose to maximize a notion of discriminatory power in the latent space. At the same time, we 
seek the latent space to minimize a notion of statistical mismatch between the source and target domains (see Fig.~\ref{fig:mainDiag} for a conceptual diagram).

\item [(iii)] Given the complexity of the resulting problem, we provide a rigorous mathematical modeling of the problem. 
In particular, we make use of the Riemannian geometry and optimization techniques on matrix manifolds to solve our learning problem\footnote{Our implementation is available on \url{https://sherath@bitbucket.org/sherath/ils.git}.}.

\item [(iv)] We extensively evaluate and contrast our solution against several baseline and state-of-the-art methods 
in addressing both unsupervised and semi-supervised DA problems.
\end{itemize}

\section{Proposed Method}
\label{sec:proposed_method}

In this work, we are interested in learning an \textbf{I}nvariant \textbf{L}atent  \textbf{S}pace (ILS) to reduce the discrepancy 
between domains.
We first define our notations.
Bold capital letters denote matrices (\eg, $\Mat{X}$) and bold lower-case letters denote column vectors (\eg, $\Vec{x}$).
$\mathbf{I}_n$ is the $n \times n$ identity matrix.
$\SPD{n}$ and $\ST{p}{n}$ denote the SPD and Stiefel manifolds, respectively, and will be formally defined later.
We show the source and target domains by $\mathcal{X}_s \subset \mathbb{R}^{s}$ and $\mathcal{X}_t \subset \mathbb{R}^{t}$.
The training samples from the source and target domains are shown by  
$\{\Vec{x}^s_i,y^s_i\}_{i = 1}^{n_s}$ and
$\{\Vec{x}^t_i\}_{i = 1}^{n_t}$, respectively. 
For now, we assume only source data is labeled. Later, we discuss how the proposed learning framework can benefit form the labeled target data.

Our idea in learning an ILS is to determine the transformations $\mathbb{R}^{s \times p} \ni \Mat{W}_s: \mathcal{X}_s \to \mathcal{H}$
and $\mathbb{R}^{t \times p} \ni \Mat{W}_t: \mathcal{X}_t \to \mathcal{H}$ from the source and target domains to a latent 
p-dimensional space $\mathcal{H} \subset \mathbb{R}^p$. We furthermore want to equip the latent space with a Mahalanobis metric, $\Mat{M} \in \SPD{p}$, to reduce the discrepancy between projected source and target samples (see Fig.~\ref{fig:mainDiag} for a conceptual diagram).

To learn $\Mat{W}_s$, $\Mat{W}_t$ and $\Mat{M}$ we propose to minimize a cost function in the form
\begin{equation}
\label{eqn:generalCost}
\mathcal{L}  = \mathcal{L}_{d} + \lambda \mathcal{L}_{u}\;.
\end{equation}
In Eq.~\ref{eqn:generalCost}, $\mathcal{L}_{d}$ is a measure of dissimilarity between labeled samples. The term
$\mathcal{L}_{u}$ quantifies a notion of  statistical difference between the source and target samples in the latent space.
As such, minimizing $\mathcal{L}$ leads to learning a latent space where not only the dissimilarity between labeled samples is reduced 
but also the domains are matched from a statistical point of view.
The combination weight $\lambda$ is envisaged  to balance the two terms. 
The subscripts ``d'' and ``u'' in Eq.~\ref{eqn:generalCost} stand for ``Discriminative''
and ``Unsupervised''. The reason behind such naming will become clear shortly.
Below we detail out the form and properties of $\mathcal{L}_{d}$ and $\mathcal{L}_{u}$.

\subsection{Discriminative Loss}

The purpose of having $\mathcal{L}_{d}$  in Eq.~\ref{eqn:generalCost} is to equip the latent space $\mathcal{H}$ with a metric
to \textbf{1.} minimize dissimilarities between samples coming from the same class and 
\textbf{2.} maximizing the dissimilarities between samples from different classes. 

Let $\Mat{Z} = \{\Vec{z}_j\}_{j = 1}^n$ be the set of labeled samples in $\mathcal{H}$.
In unsupervised domain adaptation $\Vec{z}_j = \Mat{W}_s^T\Vec{x}^s_j$ and $n = n_s$.
In the case of semi-supervised domain adaptation, 
\begin{equation*}
\Mat{Z} = \Big\{\Mat{W}_s^T\Vec{x}^s_j \Big\}_{j=1}^{n_s} \bigcup \Big\{\Mat{W}_t^T\Vec{x}^t_j \Big\}_{j=1}^{n_{tl}},
\end{equation*}
where we assume $n_{tl}$ labeled target samples are provided (out of available $n_t$ samples). 
From the labeled samples in $\mathcal{H}$, we create {$N_p$} pairs in the form 
$(\Vec{z}_{1,k},\Vec{z}_{2,k}),~k = 1,2,\cdots,{N_p}$
along their associated label $y_k \in \{-1,1\}$.
Here, $y_k = 1$ iff label of $\Vec{z}_{1,k}$ is similar to that of  $\Vec{z}_{2,k}$ 
and  $-1$ otherwise. That is the pair $(\Vec{z}_{1,k},\Vec{z}_{2,k})$ is similar if $y_k = 1$
and dissimilar otherwise.

\begin{figure}[h!]
  \centering
  \includegraphics[width=0.7\textwidth]{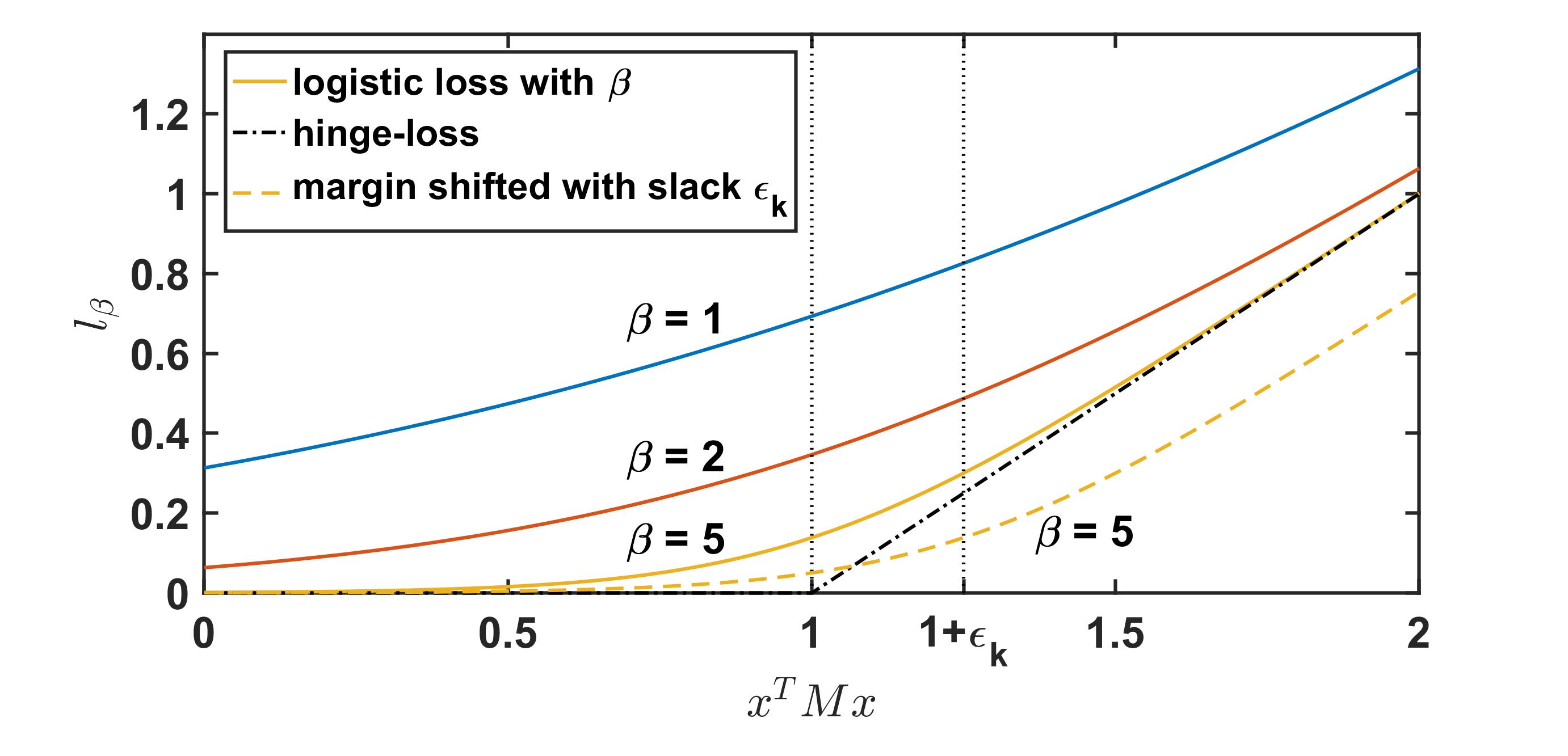}
 \caption{The behavior of the proposed $\ell_\beta$(~\ref{eqn:ell_b}) with $u=1$ for various values of parameter $\beta$. Here, 
 the horizontal axis is the value of the Mahalanobis distance and the function is plotted for $y = +1$. When $\beta\rightarrow\infty$, the function approaches the hinge-loss. An example of the soft-margin case (see Eq.~\ref{eqn:soft_margin_cost}), 
 is also plotted for $\beta=5$ curve. The figure is best seen in color.}
  \label{fig:logisticLoss}
 \end{figure}

To learn the metric $\Mat{M}$, we deem the distances between the similar pairs 
to be small while simultaneously making the distances between the dissimilar pairs large. In particular, we define $\mathcal{L}_{d}$ as,
\begin{align}
\label{eqn:distCost}
\mathcal{L}_{d} &= \frac{1}{N_p} \sum\limits_{k=1}^{N_p} \ell_\beta\big(\Mat{M},y_k,\Vec{z}_{1,k} - \Vec{z}_{2,k},1 \big)  + r(\Mat{M}),
\end{align}
with 
\begin{align}
\label{eqn:ell_b}
\hspace{-2ex} \ell_\beta\big(\Mat{M},y,\Vec{x},u \big)  = \frac{1}{\beta}\log\Big(1+\exp\big(\beta y(\Vec{x}^T\Mat{M}\Vec{x} -u )\big)\Big).
\end{align}
Here, $\ell_\beta$ is the generalized logistic function  tailored with large margin structure (see Fig.~\ref{fig:logisticLoss}) having a margin of $u$\footnote{For now we keep the margin at $u=1$ and later will use this to explain the soft-margin extension.}.
First note that the quadratic term in Eq.~\ref{eqn:ell_b} (\ie, $\Vec{x}^T\Mat{M}\Vec{x}$) measures the Mahalanobis distance between 
$\Vec{z}_{1,k}$ and $\Vec{z}_{2,k}$ if used according to Eq.~\ref{eqn:distCost}. Also note that 
$\ell_\beta\big(\cdot,\cdot,\Vec{x},\cdot \big) = \ell_\beta\big(\cdot,\cdot,-\Vec{x},\cdot \big)$, hence how samples  are order in the pairs is not important.

To better understand the behavior of the function $\ell_\beta$, assume the function is fed with a similar pair, \ie $y_k = 1$.
For the sake of discussion, also assume $\beta = 1$. In this case, $\ell_\beta$ is decreased
if the distance between $\Vec{z}_{1,k}$ and $\Vec{z}_{2,k}$ is reduced. For a dissimilar pair (\ie, $y_k = -1$), the opposite should happen to
have a smaller objective. That is, the Mahalanobis distance between the samples of a pair should be increased.

The function $\ell_\beta\big(\cdot,\cdot,\Vec{x},\cdot \big)$ can be understood as a smooth and differentiable form of the hinge-loss function. 
In fact, $\ell_\beta\big(\cdot,\cdot,\Vec{x},\cdot \big)$ asymptotically reaches the hinge-loss function  if $\beta\rightarrow\infty$. 
The smooth behavior of $\ell_\beta\big(\cdot,\cdot,\Vec{x},\cdot \big)$ is not only welcomed in the optimization scheme 
but also avoids samples in the latent space to collapse into a single point.

Along the general practice in metric learning, we regularize the metric $\Mat{M}$ by $r(\Mat{M})$. 
The divergences derived from the $\log\det(\cdot)$ function are familiar faces for regularizing Mahalanobis metrics 
in the litrature~\citep{Davis_ICML07,Saenko_ECCV10}.

Among possible choices, we make use of the Stein divergence~\citep{Anoop_TPAMI13} in this work. Hence,
\begin{equation}
r(\Mat{M}) = \frac{1}{p}\delta_s(\Mat{M},\Mat{I_p}).
\end{equation}
\noindent Where,
\begin{equation}
\delta_s(\Mat{P},\Mat{Q}) = \log\det\Big(\frac{\Mat{P}+\Mat{Q}}{2}\Big) - \frac{1}{2}\log\det \big(\Mat{P}\Mat{Q}\big),
\end{equation}
for $\Mat{P},\Mat{Q} \in \SPD{}$.

Our motivation in using the Stein divergence stems from its unique properties. 
Among others, Stein divergence is symmetric, invariant to affine transformation and closely related to geodesic distances on the 
SPD manifold~\citep{Anoop_TPAMI13,Harandi_SPD16,Cherrian_LogDet16}.

\subsection*{Soft Margin Extension}

For large values of $\beta$, the cost in Eq.~\ref{eqn:distCost} seeks the distances of similar pairs to be less than 1 while
simultaneously it deems the distances of dissimilar pairs to exceed 1. This hard-margin in the design of 
$\ell_\beta\big(\cdot,\cdot,\Vec{x},\cdot \big)$ is not desirable. 
For example, with a large number of pairs, it is often the case to have outliers. 
Forcing outliers to fit into the hard margins can result in  overfitting.
As such, we propose a soft-margin extension of Eq.~\ref{eqn:ell_b}.   
The soft-margins are implemented by associating a non-negative slack variable $\epsilon_k$ to a pair according to
\begin{align}
\label{eqn:soft_margin_cost}
\mathcal{L}_{d} &= \frac{1}{N_p} \sum\limits_{k=1}^{N_p} 
\ell_\beta\big(\Mat{M},y_k,\Vec{z}_{1,k} - \Vec{z}_{2,k},1+y_k\epsilon_k \big)  +
r(\Mat{M}) + \frac{1}{N_p} \sqrt{\sum \epsilon_k^2},
\end{align}
where a regularizer  on the slack variables is also envisaged.

\subsection*{Matching Statistical Properties}

A form of incompatibility between domains is due to their statistical discrepancies. 
Matching the first order statistics of two domains for the purpose of adaptation 
is studied in~\cite{Pan_TNN11,Masha_ICCV14,Tsai_CVPR16}\footnote{ 
The use of Maximum Mean Discrepancy (MMD)~\citep{Borgwardt_Bio06}
for domain adaptation is a well-practiced idea in the literature (see for example~\cite{Pan_TNN11,Masha_ICCV14,Tsai_CVPR16}).
Empirically, determining MMD boils down to computing the distance between domain averages when domain samples are 
lifted to a reproducing kernel Hilbert space. 
Some studies claim matching the first order statistics is a weaker form of 
domain adaptation through MMD. We do not support this claim and hence do not see our solution as 
a domain adaptation method by minimizing the MMD.}.
In our framework, matching domain averages can be achieved readily. In particular, 
let $\bar{\Vec{x}}^s_i = \Vec{x}^s_i - \Vec{m}_s$ and $\bar{\Vec{x}}^t_j = \Vec{x}^t_j - \Vec{m}_t$ be the centered 
source and target samples  with $\Vec{m}_s$ and $\Vec{m}_t$ being the mean of corresponding domains. 
It follows easily that the domain means in the latent space are zero\footnote{We note that  
$\sum \Mat{W}_s^T\bar{\Vec{x}}^s_i = \Mat{W}_s^T\sum\bar{\Vec{x}}^s_i = \Vec{0}$. This holds for the target domain as well.} and hence matching is achieved.

To go beyond first order statistics, we propose to match the second order statistics (\ie, covariance matrices)
as well.
The covariance of a domain reflects the relationships between its features. 
Hence, matching covariances of source and target domains in effect improves the cross feature relationships. 
We capture the mismatch between source and target covariances in the latent space using the $\mathcal{L}_{u}$ loss 
in Eq.~\ref{eqn:generalCost}. Given the fact that covariance matrices are points on the SPD manifold, we make use of the Stein divergence to measure their differences. This leads us to define $\mathcal{L}_{u}$ as  
\begin{equation}
\label{eqn:statCost}
\begin{split}
\mathcal{L}_{u}  = \frac{1}{p}\delta_s(\Mat{W}_s^T \Mat{\Sigma}_s\Mat{W}_s,\Mat{W}_t^T\Mat{\Sigma}_{t}\Mat{W}_t),
\end{split}
\end{equation}
with $\Mat{\Sigma}_s \in \SPD{s}$ and $\Mat{\Sigma}_t \in \SPD{t}$ being the covariance matrices of the source and target domains, respectively. We emphasize that  matching the statistical properties as discussed above is an unsupervised technique, enabling us to address unsupervised DA. 


\subsection{Classification Protocol}

Upon learning $\Mat{W}_s$, $\Mat{W}_t$, $\Mat{M}$, training samples from the source and target (if available) domains are mapped 
to the latent space using $\Mat{W}_s\Mat{M}^{\frac{1}{2}}$ and $\Mat{W}_t\Mat{M}^{\frac{1}{2}}$, respectively. 
For a query from the target domain $\Vec{x}_q^t$, $\Mat{M}^{\frac{1}{2}}\Mat{W}_t^T\Vec{x}_q^t$  
is its latent space representation which is subsequently classified by a nearest neighbor classifier.



\section{Optimization}
\label{sec:opt}

The objective of our algorithm is to learn the transformation parameters ($\Mat{W}_s$ and $\Mat{W}_t$), the metric $\Mat{M}$ and slack variables  $\epsilon_1,\epsilon_2,...\epsilon_{N_p}$ (see Eq.~\ref{eqn:soft_margin_cost} and Eq.~\ref{eqn:statCost}).  
Inline with the general practice of dimensionality reduction, we propose to have orthogonality constraints on $\Mat{W}_s$ and $\Mat{W}_t$. 
That is $\Mat{W}_s^T\Mat{W}_s = \Mat{W}_t^T\Mat{W}_t = \mathbf{I}_p$. We elaborate how orthogonality 
constraint improves the discriminatory power of the proposed framework later in our experiments.

The problem  depicted in Eq.~\ref{eqn:generalCost} is indeed a non-convex and constrained optimization problem. 
One may resort to the method of Projected Gradient Descent (PGD)~\citep{Boyd_Convex04} to minimize Eq.~\ref{eqn:generalCost}. 
In PGD, optimization is proceed by projecting the gradient-descent updates onto the set of constraints. For example, in 
our case, we can first update $\Mat{W}_s$ by ignoring the orthogonality constraint on $\Mat{W}_s$ and then project the result onto
the set of orthogonal matrices 
using eigen-decomposition. As such, optimization can be performed by alternatingly updating 
$\Mat{W}_s$, $\Mat{W}_t$, the metric $\Mat{M}$ and slack variables using PGD.

In PGD, to perform the projection, the set of constraints needs to be closed though in practice one can resort to 
open sets. For example, the set of SPD matrices is open though one can project a symmetric matrix onto this set using eigen-decomposition. 

Empirically, PGD showed an erratic and numerically unstable  behavior in addressing our problem. 
This can be attributed to the non-linear nature of Eq.~\ref{eqn:generalCost}, existence of open-set constraints in the problem or perhaps 
the combination of both.
To alleviate the aforementioned difficulty, we propose a more principle approach to minimize Eq.~\ref{eqn:generalCost} by making use of the Riemannian optimization technique. We take a short detour and briefly describe the Riemannian optimization methods below.

\subsection*{Optimization on Riemannian manifolds.}
Consider  a non-convex constrained problem in the form 
\begin{align}
\mathrm{minimize}~f(\Vec{x}) \notag\\
\subject~~~\Vec{x} \in \mathcal{M}\;,
\label{eqn:opt_riemannian_manifold}
\end{align}
where $\mathcal{M}$ is a Riemannian manifold, \ie, informally,  a smooth surface that locally resembles a Euclidean space.
Optimization techniques on Riemannian manifolds (\eg, Conjugate Gradient)
start with an initial solution $\Vec{x}^{(0)} \in \mathcal{M}$, and iteratively improve the 
solution by following the geodesic identified by the gradient. For example, in the case of Riemannian Gradient Descent Method (RGDM),  
the updating rule reads
\begin{equation}
\Vec{x}^{(t+1)} = \tau_{\Vec{x}^{(t)}} \big( -\alpha~\mathrm{grad}~f(\Vec{x}^{(t)})\big)\;,
\end{equation}
with $\alpha > 0$ being the algorithm's step size. Here, $\tau_{\Vec{x}} ( \cdot): T_\Vec{x}\mathcal{M} \to \mathcal{M}$, 
is called the retraction\footnote{Strictly speaking and in contrast with the exponential map, a retraction only guarantees to pull a tangent vector on the geodesic locally, \ie, close to the origin of the tangent space. Retractions, however, are typically easier to compute than the exponential map and have proven effective in Riemannian optimization~\citep{Absil_Book}.} and moves the solution along the descent direction while assuring that the new solution is on the 
manifold $\mathcal{M}$, \ie, it is within the constraint set. 
$T_\Vec{x}\mathcal{M}$ is the tangent space of $\mathcal{M}$ at $\Vec{x}$ and can be thought of as a vector space 
with its vectors being the gradients of all functions defined on $\mathcal{M}$. 

We defer more details on Riemannian optimization techniques to the \textbf{Appendix A}. 
As for now, it suffices to say that to perform optimization on the Riemannian manifolds, 
the form of Riemannian gradient, retraction and the gradient of the objective with respect to its parameters 
(shown by $\nabla$) are required. 
The constraints in Eq.\ref{eqn:generalCost} are orthogonality (transformations $\Mat{W}_s$ and $\Mat{W}_t$) and p.d. for metric $\Mat{M}$.
The geometry of these constraints are captured by the Stiefel~\citep{James_Stiefel76,Harandi_GB16} and SPD~\citep{Harandi_SPD16,Cherrian_SPD2016} manifolds, formally defined as

\begin{definition}[The Stiefel Manifold]
The set of ($n \times p$)-dimensional matrices, $p \leq n$, with orthonormal columns endowed with the Frobenius inner product\footnote{%
Note that the literature is divided between this choice and another form of Riemannian metric. 
See~\cite{Edelman_1998} for details.} forms a compact Riemannian manifold
called the  Stiefel manifold $\ST{n}{p}$~\citep{Absil_Book}.
\begin{equation}
\ST{n}{p} \triangleq \{\Mat{W} \in \mathbb{R}^{n \times p}: \Mat{W}^T\Mat{W} = \mathbf{I}_p\}\;.
\label{eqn:stiefel_manifold}
\end{equation}
\end{definition}

\begin{definition}[The SPD Manifold]
The set of ($p \times p$) dimensional real, SPD matrices endowed with the Affine Invariant Riemannian Metric (AIRM)~\citep{Pennec_IJCV06} forms the  SPD manifold $\SPD{p}$. 
\begin{equation}
\hspace{-0.5ex}\SPD{p} \triangleq \{\Mat{M} \in \mathbb{R}^{p \times p} \hspace{-0.5ex}: \Vec{v}^T\Mat{M}\Vec{v} >0,~\hspace{-0.5ex}\forall \Vec{v} \in \mathbb{R}^p-\{\Vec{0}_p\}\}.
\label{eqn:spd_manifold}
\end{equation}
\end{definition}

Updating $\Mat{W}_s$, $\Mat{W}_t$ and $\Mat{M}$ and slacks can be done alternatively using Riemannian optimization. 
As mentioned above, the ingredients for doing so are \textbf{1.} the Riemannian tools for the Stiefel and SPD manifolds 
along \textbf{2.} the form of gradients of the objective with respect to its parameters. To do complete justice, 
in Table.~\ref{tab:ReimannianGrads} we provide the Riemmanian metric, form of Riemannian gradient and retraction for the Stiefel and SPD manifolds.
In Table.~\ref{tab:gradientsComputed}, the gradient of Eq.~\ref{eqn:generalCost} with respect to $\Mat{W}_s$, $\Mat{W}_t$ and $\Mat{M}$ and slacks is provided.
The detail of derivations can be found in the \textbf{Appendix B}. 
A tiny note about the slacks worth mentioning. To preserve the non-negativity constraint on  $\epsilon_k$, we define 
$\epsilon_k = e^{v_k}$ and optimize on $v_k$ instead. This in turn makes optimization for the slacks an unconstrained problem.

\begin{remark}
From a geometrical point of view, we can make use of the product topology of the parameter space to avoid alternative optimization. More specifically, the set 
\begin{equation}
\label{e:prodTopology}
\mathcal{M}_{prod.} = \ST{s}{p} \times \ST{t}{p} \times \SPD{p} \times \mathbb{R}^{N_p},
\end{equation} 
can be given the structure of a Riemannian manifold using the concept of product topology~\citep{Absil_Book}. 
\end{remark}

\begin{figure}[t!]
 \centering
  \includegraphics[width=0.52\textwidth]{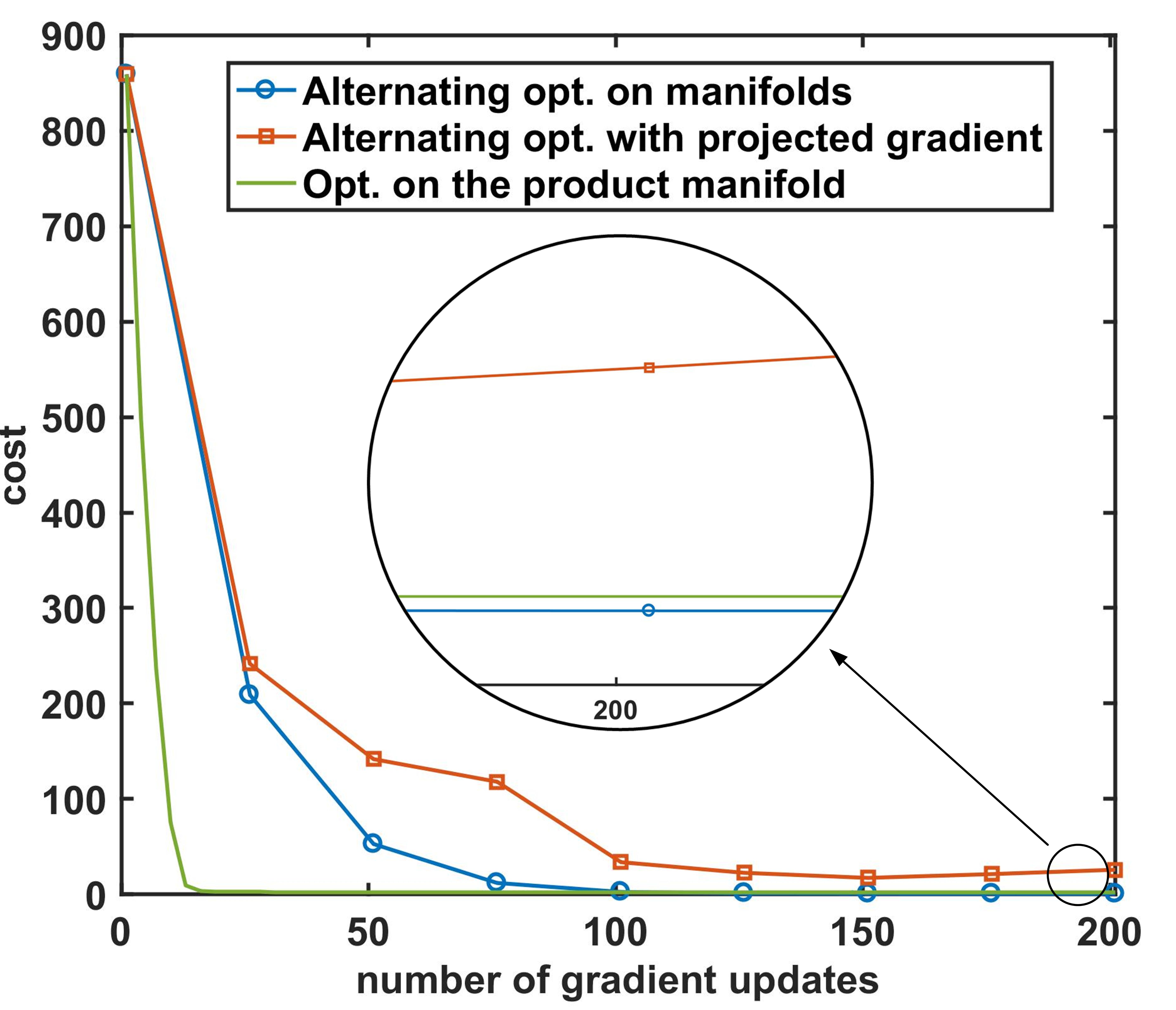}
 \caption{Optimizing Eq.~\ref{eqn:generalCost} using PGD (red curve), Riemannian gradient descent using alternating approach (blue curve)
 and product topology (green curve). Optimization using the product topology converges faster but a lower cost can be attained using 
 alternating Riemannian optimization. }
 \label{fig:OptimizationCost}
 \end{figure}

\begin{remark}
In Fig.~\ref{fig:OptimizationCost}, we compare the convergence behavior of PGD, alternating Riemannian optimization and 
optimization using the product geometry. While optimization on $\mathcal{M}_{prod.}$ convergences faster, the alternating method  
results in a lower loss. This behavior resembles the difference 
between the stochastic gradient descent compared to its batch counterpart. 
\end{remark}

\begin{remark}
The complexity of the optimization depends on the number of labeled pairs. One can always resort to a stochastic
solution~\cite{Song_CVPR16,Sa_ICML15,Bonnabel_Stoch13} by sampling from the set of similar/dissimilar pairs if addressing a very large-scale problem. In our experiments, we 
did not face any difficulty optimizing with an i7 desktop machine with 32GB of memory.
\end{remark}

\begin{table*}[t]
\caption{Riemannian metric, gradient and retraction on $\ST{n}{p}$ and $\SPD{p}$. Here, 
$\mathrm{uf}(\Mat{A}) = \Mat{A}(\Mat{A}^T\Mat{A})^{-1/2}$, which yields an orthogonal matrix, $\mathrm{sym}(\Mat{A}) = \frac{1}{2}(\Mat{A} + \Mat{A}^T)$ and
$\expm(\cdot)$ denotes the matrix exponential.}
\label{tab:ReimannianGrads}
\small
\centering
\begin{tabular}{|c|c|c|}
\hline
{\bf } &{\bf $\ST{n}{p}$} &{\bf $\SPD{p}$}\\
\hline 
Matrix representation 					&$\Mat{W} \in \mathbb{R}^{n \times p}$		&$\Mat{M} \in \mathbb{R}^{p \times p}$\\
Riemannian metric						&$g_\nu(\xi,\varsigma) = \tr(\xi^T\varsigma)$
										&$g_\mathcal{S}(\xi,\varsigma) =	\tr \left( \Mat{M}^{-1} \xi \Mat{M}^{-1}\varsigma\right)$\\
Riemannian gradient						&$\nabla_\Mat{W}(f)  - \Mat{W}\mathrm{sym}\Big(\Mat{W}^T \nabla_\Mat{W}(f)\Big)$
										&$\Mat{M}\mathrm{sym}\Big(\nabla_\Mat{M}(f)\Big)\Mat{M}$\\
Retraction								&$\mathrm{uf}(\Mat{W} + \xi)$
										&$\Mat{M}^\frac{1}{2}\expm(\Mat{M}^{-\frac{1}{2}}\xi\Mat{M}^{-\frac{1}{2}})\Mat{M}^\frac{1}{2}$
								\\
\hline
\end{tabular}
\label{tab:riemannian_tools}
\end{table*}

\begin{table*}[]
\centering
\caption{Gradients of soft-margin $\ell_\beta$ and $\mathcal{L}_{u}$ w.r.t. the model parameters and slack variables. Without less of generality we only consider a labeled similar ($y_k = +1$)  pair $\Vec{x}^s_i$ and $\Vec{x}^t_j$. Here, $r = \exp{\big(\beta \big((\Mat{W}_s^T\Vec{x}^s_i - \Mat{W}_t^T\Vec{x}^t_i)^T\Mat{M}(\Mat{W}_s^T\Vec{x}^s_i - \Mat{W}_t^T\Vec{x}^t_i) - 1 -e^{v_k}\big)}\big)$.}
\label{tab:gradientsComputed}
\scalebox{1}{
\begin{tabular}{|c|l|}
\hline
$ \nabla_{\Mat{W}_s}\ell_\beta $ & 
$\frac{2}{N_p}(1 + r^{-1})^{-1}\Vec{x}^s_i({\Vec{x}^s_i}^T\Mat{W}_s - {\Vec{x}^t_j}^T\Mat{W}_t)\Mat{M} 
$ \\ \hline
$\nabla_{\Mat{W}_t}\ell_\beta$  & 
$\frac{2}{N_p}(1 + r^{-1})^{-1}\Vec{x}^t_j({\Vec{x}^t_j}^T\Mat{W}_t - {\Vec{x}^s_i}^T\Mat{W}_s)\Mat{M} 
$ \\ \hline
$\nabla_{\Mat{M}}\ell_\beta$ & $\frac{1}{N_p}(1 + r^{-1})^{-1}\big(\Mat{W}_s^T{\Vec{x}^s_i}-\Mat{W}_t^T\Vec{x}^t_j\big)
\big({\Vec{x}^s_i}^T\Mat{W}_s - {\Vec{x}^t_j}^T\Mat{W}_t\big)$ \\ \hline
$\nabla_{v_k}\ell_\beta$ & $\frac{-1}{N_p}e^{v_k}(1 + r^{-1})^{-1}$ \\ \hline
$\nabla_{\Mat{W}_s}\mathcal{L}_{u}$  & 
$ \frac{1}{p}\Mat{\Sigma}_s\Mat{W}_s \Big(2\big(\Mat{W}_s^T\Mat{\Sigma}_s\Mat{W}_s + \Mat{W}_t^T\Mat{\Sigma}_t\Mat{W}_t\big)^{-1}  
- \big( \Mat{W}_s^T\Mat{\Sigma}_s\Mat{W}_s \big)^{-1}\Big)$ \\ \hline
\end{tabular}}
\end{table*}



\section{Related Work}
\label{sec:related_work}
The literature on domain adaptation spans a very broad range (see~\cite{Patel_Survey15} for a recent survey). 
Our solution falls under the category of domain adaptation by subspace learning (DA-SL). As such, 
we confine our review only to methods under the umbrella of  DA-SL.

One notable example of constructing a latent space is the work of~\cite{Daume_DA10}.
In particular, the authors propose to use two fixed and predefined transformations to project source and 
target data to a common and higher-dimensional space. As a  requirement, the method only accepts 
domains with the same dimensionality and hence cannot be directly used to adapt heterogeneous domains.

\cite{Gopalan_ICCV11} observed that the geodesic connecting the source and target subspaces conveys useful information for 
DA and proposed the Sampling Geodesic Flow (SGF) method~\citep{Gopalan_ICCV11}.
The Geodesic Flow Kernel (GFK)  is an 
improvement over the SGF technique where instead of sampling a few points 
on the geodesic, the whole curve is used for domain adaptation~\citep{Gong_CVPR12}. 
In both methods, the domain subspaces are fixed and obtained by Principal Component Analysis (PCA) or 
Partial Least Square regression (PLS)~\citep{Krishnan_PLS11}. 
In contrast to our solution, in SGF and GFK learning the domain subspaces is disjoint from the knowledge transfer algorithm. 
In our experiments, we will see that the subspaces determined by our method can even boost the performance of GFK, 
showing the importance of  joint learning of domain subspaces and the knowledge transfer scheme. In~\cite{Ni_Subspace13} dictionary learning is used for interpolating the intermediate subspaces.

Domain adaptation by fixing the subspace/representation of one of the domains is a popular theme 
in many recent works, as it simplifies the learning scheme. Examples are 
the max-margin adaptation~\citep{Hoffman_IJCV14,Duan_ICML12}, 
the metric/similarity learning of~\cite{Saenko_ECCV10} and its kernel extension~\citep{Kulis_CVPR11},
the landmark approach of~\cite{Tsai_CVPR16}, the alignment 
technique of~\cite{Basura_ICCV13,Fernando_PRL15}, correlation matching of~\cite{SUN_Return15} 
and methods that use maximum mean discrepancy (MMD)~\citep{Borgwardt_Bio06} for domain adaptation~\citep{Pan_TNN11,Masha_ICCV14}.  

In contrast to the above methods, some studies opt to learn the domain representation along the knowledge transfer method jointly.
Two representative works are the  HeMap~\citep{Shi_ICDM10} and manifold alignment~\citep{Wang_IJCAI11}.
The HeMap  learns two projections to minimize the instance discrepancies~\citep{Shi_ICDM10}. 
The problem is however formulated such that equal number of source and target instances is required to perform the training. 
The manifold alignment algorithm of~\cite{Wang_IJCAI11} attempts to preserve the label structure in the latent space. However, it is essential for the algorithm to have access to labeled data in both source and target domains.

Our solution learns all transformations to the latent space. We do not resort at subspace representations learned disjointly to the DA framework. With this use of the latent space, our algorithm is not limited for applications where source and target data have similar dimensions or structure.



\section{Experimental Evaluations}
\label{sec:experiments}

We run extensive experiments on both semi-supervised and unsupervised settings, spanning from the handcrafted features (SURF) to the current state-of-the-art deep-net features (VGG-Net).
 For comparisons, we use the implementations made available by the original authors. Our method is denoted by \textbf{ILS}. 

\subsection{Implementation Details}

Since the number of dissimilar pairs is naturally larger than the number of similar pairs, we randomly sample from the different pairs to keep the sizes of these two sets equal. 
We initialize the projection matrices $\Mat{W_s}$, $\Mat{W_t}$ with PCA, following the transductive protocol~\cite{Gong_CVPR12, Basura_ICCV13, Hoffman_IJCV14, Tsai_CVPR16}. For the semi-supervised setting, we initialize $\Mat{M}$ with the Mahalanobis metric learned on the similar pair covariances~\citep{Koestinger_CVPR12}, and for the unsupervised setting, we initialize it with the identity matrix. For all our experiments we have $\lambda = 1$. We include an experiment showing our solution's robustness to $\lambda$ in the supplementary material. We use the toolbox provided by~\cite{manopt} for our implementations. 

\begin{remark}
To have a simple way of determining $\beta$ in Eq.~\ref{eqn:ell_b}, we propose a heuristic which is shown to be effective in our experiments.
To this end, we propose to set $\beta$ to the reciprocal of the standard deviation of the similar pair distances.
\end{remark}

\subsection{Semi-supervised Setting}

 \begin{figure}[t!]
 \centering
  \includegraphics[width=0.8\textwidth]{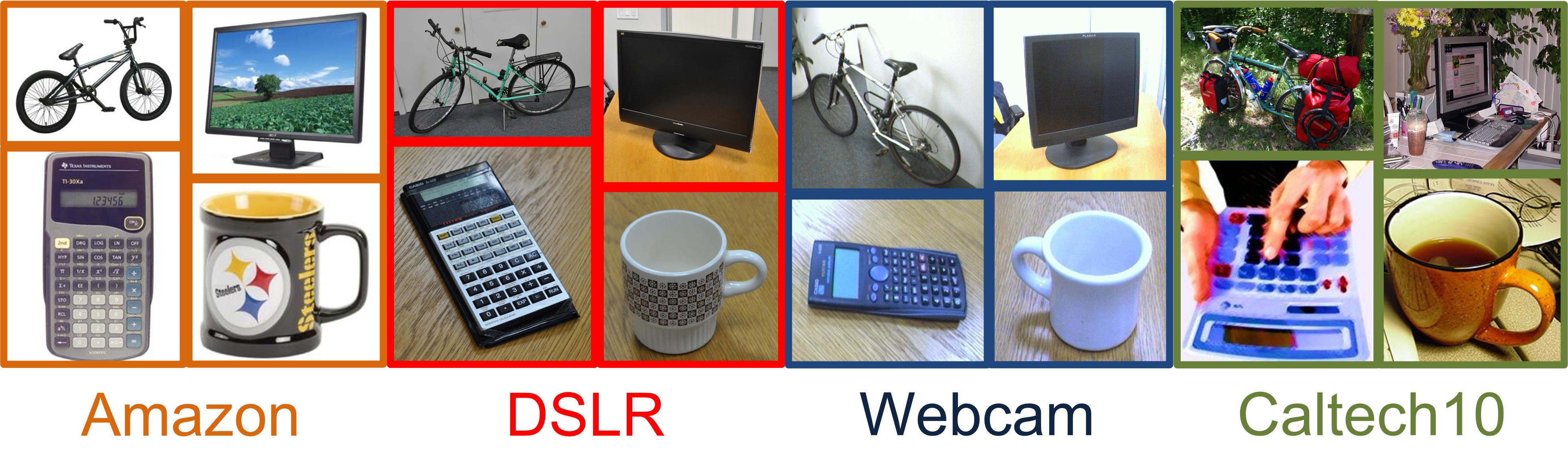}
 \caption{Samples from the Office$+$Caltech10 dataset~\citep{Saenko_ECCV10, Gong_CVPR12}. Although the DSLR and Webcam images depict similar content, they have considerably different resolutions.} 
 \label{fig:AWDC}
\end{figure}

In our semi-supervised experiments, we follow the standard setup on the Office$+$Caltech10 
dataset with the train/test splits provided by~\cite{Hoffman_ICLR2013}. The Office+Caltech10 dataset contains images collected from 4 different sources (see Fig.~\ref{fig:AWDC}) and 10 object classes. The corresponding domains are \textbf{A}mazon, \textbf{W}ebcam, \textbf{D}SLR, and \textbf{C}altech. We use a subspace of dimension 20 for DA-SL algorithms. 
We employ SURF~\citep{Bay_SURF06} for the handcrafted feature experiments. We extract VGG-Net features with the network model of~\citep{Simonyan_VGG14} for the deep-net feature experiments\footnote{The same SURF and VGG-Net features are used for the unsupervised experiments as well.}. We compare our performance with the following benchmarks:

\noindent\textbf{1-NN-t and SVM-t :} Basic Nearest Neighbor (1-NN) and linear SVM classifiers trained only on the target domain.

\noindent\textbf{HFA~\citep{Duan_ICML12} :} This method employs latent space learning based on the max-margin framework. As in its original implementation, we use the RBF kernel SVM for its evaluation.

\noindent\textbf{MMDT~\citep{Hoffman_IJCV14} :} This method jointly learns a transformation between the source and target domains along a linear SVM for classification.

\noindent\textbf{CDLS~\citep{Tsai_CVPR16} :} This is the cross-domain landmark search algorithm. We use the parameter setting ($\delta = 0.5$ in the notation of~\cite{Tsai_CVPR16}) recommended by the authors. 

Table~\ref{tab:SemiSuper_SURF}, Table~\ref{tab:SemiSuper_VGGFC6} and  Table~\ref{tab:Super_FC7} report the performances using the handcrafted SURF, VGG-FC6 and VGG-FC7 features, respectively. For the SURF features our solution achieves the best performance in 7 out 12 cases. For the VGG-FC6  and VGG-FC7 features, our solution tops in 9 and 7 sets respectively. It seems that in comparison to  VGG-FC6, VGG-FC7 features are less discriminative for all the DA algorithms.  We notice the \textbf{1-NN-t} baseline performs the worst for both SURF  and the VGG-Net features. Hence, it is clear that the used features do not favor the nearest neighbor classifier. We observe that \textbf{C}altech and \textbf{A}mazon domains contain the largest number of test instances. Although the performances of all tested methods decrease on these domains, particularly on \textbf{C}altech, our method achieves the top rank in almost all domain transformations.

\begin{table*}[!t]
\centering
\small
\caption{Semi-supervised domain adaptation results using SURF features on Office$+$Caltech10~\cite{Gong_CVPR12} dataset with the evaluation setup of~\cite{Hoffman_IJCV14}. The best score (in bold blue), the second best (in blue). 
}
\label{tab:SemiSuper_SURF}
\scalebox{0.85}{
\begin{tabular}{|r|c|c|c|c|c|c|c|c|c|c|c|c|}
\hline
      & A$\rightarrow$W   & A$\rightarrow$D   & A$\rightarrow$C   & W$\rightarrow$A   & W$\rightarrow$D   & W$\rightarrow$C   & D$\rightarrow$A   & D$\rightarrow$W   & D$\rightarrow$C   & C$\rightarrow$A   & C$\rightarrow$W   & C$\rightarrow$D    \\ \hline
1-NN-t    & 34.5 & 33.6 & 19.7  & 29.5  & 35.9  &  18.9 & 27.1  &  33.4 & 18.6  & 29.2  & 33.5  & 34.1    \\      
SVM-t & 63.7 & {\color{Blue}57.2} & 32.2 & 46.0 & 56.5 & 29.7 & 45.3 & 62.1 & 32.0 & 45.1 & 60.2 & 56.3    \\
HFA  & 57.4 & 55.1 & 31.0 & \textbf{\color{blue}56.5} & 56.5 & 29.0 & 42.9 & 60.5 & 30.9 & 43.8 & 58.1 & 55.6  \\
MMDT & {\color{Blue}64.6} & 56.7 & {\color{Blue}36.4} & 47.7 & {\color{Blue}67.0} & 32.2 & 46.9 & {\color{Blue}74.1} & 34.1 & 49.4 & {\color{Blue}63.8} & {\color{Blue}56.5}   \\
CDLS & \textbf{\color{blue}68.7} & \textbf{\color{blue}60.4} & 35.3 & 51.8 & 60.7 & {\color{Blue}33.5} & {\color{Blue}50.7} & 68.5 & {\color{Blue}34.9} & {\color{Blue}50.9} & \textbf{\color{blue}66.3} & \textbf{\color{blue}59.8}   \\
\textbf{ILS} (1-NN)  & 59.7 & 49.8 & \textbf{\color{blue}43.6} & {\color{Blue}54.3} & \textbf{\color{blue}70.8} & \textbf{\color{blue}38.6} & \textbf{\color{blue}55.0} & \textbf{\color{blue}80.1} & \textbf{\color{blue}41.0} & \textbf{\color{blue}55.1} & 62.9 & 56.2    \\ \hline
\end{tabular}}
\vspace{-1mm}
\end{table*}

\begin{table*}[!t]
\centering
\small
\caption{Semi-supervised domain adaptation results using VGG-FC6 features on Office$+$Caltech10~\cite{Gong_CVPR12} dataset with the evaluation setup of~\cite{Hoffman_IJCV14}. The best (in bold blue), the second best (in blue).}
\label{tab:SemiSuper_VGGFC6}
\scalebox{0.85}{
\begin{tabular}{|r|c|c|c|c|c|c|c|c|c|c|c|c|}
\hline
      & A$\rightarrow$W   & A$\rightarrow$D   & A$\rightarrow$C   & W$\rightarrow$A   & W$\rightarrow$D   & W$\rightarrow$C   & D$\rightarrow$A   & D$\rightarrow$W   & D$\rightarrow$C   & C$\rightarrow$A   & C$\rightarrow$W   & C$\rightarrow$D   \\ \hline
1-NN-t    & 81.0  & 79.1  &  67.8  &  76.1  &  77.9  &  65.2  &  77.1  &  81.7  &  65.6  &  78.3  & 80.2  &  77.7    \\      
SVM-t     & 89.1  & \textbf{\color{blue}88.2}  & 77.3  &  86.5  &  87.7 &  76.3  & 87.3  & 88.3  & 76.3  & 87.5  &  87.8 & 84.9   \\
HFA &  87.9 & 87.1  & 75.5  &  85.1  & 87.3  &  74.4  &  85.9 & 86.9  &  74.8 & 86.2  & 86.0  & \textbf{\color{blue}87.0} \\
MMDT & 82.5  & 77.1  & {\color{Blue}78.7}  &  84.7  &  85.1 &  73.6  & 83.6  & 86.1  &  71.8 & 85.9  & 82.8  & 77.9   \\
CDLS &  \textbf{\color{blue}91.2} & 86.9  & 78.1  &  {\color{Blue}87.4}  & {\color{Blue}88.5} &  {\color{Blue}78.2}  & {\color{Blue}88.1}  &  {\color{Blue}90.7} & {\color{Blue}77.9}  & {\color{Blue}88.0}  & {\color{Blue}89.7}  & 86.3   \\
\textbf{ILS} (1-NN)  & {\color{Blue}90.7}  & {\color{Blue}87.7}  &  \textbf{\color{blue}83.3} & \textbf{\color{blue}88.8} & \textbf{\color{blue}94.5}  &  \textbf{\color{blue}82.8} & \textbf{\color{blue}88.7} & \textbf{\color{blue}95.5}  &  \textbf{\color{blue}81.4} & \textbf{\color{blue}89.7}  & \textbf{\color{blue}91.4}  &  {\color{Blue}86.9}     \\ \hline
\end{tabular}}
\vspace{-1mm}
\end{table*}

\smallskip
\begin{table*}[!t]
\small
\centering
\caption{Semi-supervised domain adaptation results using VGG-FC7 features on Office$+$Caltech10~\cite{Gong_CVPR12} dataset with the evaluation setup of~\cite{Hoffman_IJCV14}. The best (in bold blue), the second best (in blue).}
\label{tab:Super_FC7}
\scalebox{0.85}{
\begin{tabular}{|r|c|c|c|c|c|c|c|c|c|c|c|c|c|}
\hline
      & A$\rightarrow$W   & A$\rightarrow$D   & A$\rightarrow$C   & W$\rightarrow$A   & W$\rightarrow$D   & W$\rightarrow$C   & D$\rightarrow$A   & D$\rightarrow$W   & D$\rightarrow$C   & C$\rightarrow$A   & C$\rightarrow$W   & C$\rightarrow$D  \\ \hline
1-NN-t    & 81.8 & 78.2 & 68.3 & 77.8 & 77.6 & 67.4 & 78.1 & 81.5 & 66.9 & 79.0 & 80.6 & 77.4  \\
SVM-t   & 87.5 & \textbf{{\color{blue}85.4}} & 76.8 & 86.2 & 85.6 & 75.8 & {\color{blue}87.0} & 87.1 & 76.0 & 87.1 & 86.4 & 84.4 \\
HFA & 86.6 & {\color{blue}85.3} & 75.2 & 84.9 & 85.5 & 74.8 & 85.8 & 86.5 & 75.1 & 86.0 & 85.3 & \textbf{{\color{blue}84.8}} \\
MMDT & 76.9 & 73.3 & 78.1 & 83.6 & 79.5 & 72.2 & 82.3 & 83.8 & 71.7 & 85.3 & 77.8 & 72.6  \\
CDLS  & {\color{blue}\textbf{90.0}} & 85.0 & {\color{blue}78.5} & {\color{blue}87.2} & {\color{blue}86.5} & {\color{blue}79.0} & \textbf{{\color{blue}87.7}} & {\color{blue}89.5} & \textbf{{\color{blue}78.8}} & {\color{blue}87.8} & \textbf{{\color{blue}89.7}} & {\color{blue}84.6} \\
\textbf{ILS} (1-NN)    & {\color{blue}89.3} & 84.0 & \textbf{{\color{blue}81.9}} & \textbf{{\color{blue}88.4}} & \textbf{{\color{blue}91.0}} & \textbf{{\color{blue}80.8}} & 86.9 & \textbf{{\color{blue}94.4}} & \textbf{{\color{blue}78.8}} & \textbf{{\color{blue}88.9}} & {\color{blue}88.7} & 83.3  \\ \hline
\end{tabular}}
\vspace{-1mm}
\end{table*}

\subsection{Unsupervised Setting}

In the unsupervised domain adaptation problem, only labeled data from the source domain is available~\citep{Basura_ICCV13, Gong_CVPR12}. We perform two sets of experiments for this setting. (1) We evaluate the object recognition performance on the Office+Caltech10 dataset. Similar to the semi-supervised settings, we use the SURF and VGG-Net features. Our results demonstrate that the learned transformations by our method are superior domain representations. (2) We analyze our performance when the domain discrepancy is gradually increased. This experiment is performed on the PIE-Face dataset. We compare our method with the following benchmarks: 

\smallskip
\noindent\textbf{1-NN-s and SVM-s :} Basic 1-NN and linear SVM classifiers trained only on the source domain.

\noindent \textbf{GFK-PLS~\citep{Gong_CVPR12} :} The geodesic flow kernel algorithm where  partial least squares (PLS) implementation is used to initialize the source subspace. Results are evaluated on kernel-NNs.

\noindent \textbf{SA~\citep{Basura_ICCV13} :} This is the subspace alignment algorithm. Results are evaluated using 1-NN.

\noindent \textbf{CORAL~\citep{SUN_Return15} :} The correlation alignment algorithm that uses a linear SVM on the similarity matrix formed by correlation matching.

\subsubsection{Office+Caltech10 (Unsupervised)}
We follow the original protocol provided by~\cite{Gong_CVPR12} on Office+Caltech10 dataset. 
Note that several baselines, determine the best dimensionality per domain to achieve their maximum accuracies on SURF features. 
We observed that a dimensionality in the range [20,120] provides consistent results for our solution using SURF features. For VGG features we empirically found the dimensionality of 20 suits best for the compared DA-SL algorithms. 

Table.~\ref{tab:Unsuper_SURF}, Table.~\ref{tab:Unsuper_VGG} and Table.~\ref{tab:Unsuper_FC7} present the unsupervised setting results using the SURF, VGG-FC6 and VGG-FC7 features. For all feature types, our solution yields the best performance in 8 domain transformations out of 12. Similarly to the semi-supervised experiments we notice the VGG-FC7 is less favorable for DA algorithms.

\smallskip
\begin{table*}[!t]
\small
\centering
\caption{Unsupervised domain adaptation results using SURF features on Office$+$Caltech10~\cite{Gong_CVPR12} dataset with the evaluation setup of~\cite{Gong_CVPR12}.The best (in bold blue), the second best (in blue).}
\label{tab:Unsuper_SURF}
\scalebox{0.85}{
\begin{tabular}{|r|c|c|c|c|c|c|c|c|c|c|c|c|}
\hline
      & A$\rightarrow$W   & A$\rightarrow$D   & A$\rightarrow$C   & W$\rightarrow$A   & W$\rightarrow$D   & W$\rightarrow$C   & D$\rightarrow$A   & D$\rightarrow$W   & D$\rightarrow$C   & C$\rightarrow$A   & C$\rightarrow$W   & C$\rightarrow$D   \\ \hline
1-NN-s    & 23.1 & 22.3 & 20.0 & 14.7 & 31.3 & 12.0 & 23.0 & 51.7 & 19.9 & 21.0 & 19.0 & 23.6   \\
SVM-s   & 25.6 & 33.4 & 35.9 & 30.4 & 67.7 & 23.4 & 34.6 & 70.2 & 31.2 & 43.8 & 30.5 & 40.3    \\
GFK-PLS & 35.7 & 35.1 & {\color{Blue}37.9} & 35.5 & 71.2 & 29.3 & 36.2 & 79.1 & 32.7 & 40.4 & 35.8 & {\color{Blue}41.1}  \\
SA     & 38.6 & 37.6 & 35.3 & 37.4 & {\color{Blue}80.3} & 32.3 & 38.0 & {\color{Blue}83.6} & 32.4 & 39.0 & 36.8 & 39.6    \\
CORAL   & {\color{Blue}38.7} & {\color{Blue}38.3} & \textbf{\color{blue}40.3} & {\color{Blue}37.8} & \textbf{\color{blue}84.9} & \textbf{\color{blue}34.6} & {\color{Blue}38.1} & \textbf{\color{blue}85.9} & {\color{Blue}34.2} & {\color{Blue}47.2} & {\color{Blue}39.2} & 40.7     \\
\textbf{ILS} (1-NN)   & \textbf{\color{blue}40.6} &   \textbf{\color{blue}41.0}   &   37.1   &   \textbf{\color{blue}38.6}   &   72.4   &  {\color{Blue}32.6}    &  \textbf{\color{blue}38.9}    &  79.1  &  \textbf{\color{blue}36.9}  &   \textbf{\color{blue}48.6}   &  \textbf{\color{blue}42.0}    & \textbf{\color{blue}44.1}   \\ \hline
\end{tabular}}
\vspace{-1mm}
\end{table*}

\begin{table*}[!t]
\centering
\small
\caption{Unsupervised domain adaptation results using VGG-FC6 features on Office$+$Caltech10~\cite{Gong_CVPR12} dataset with the evaluation setup of~\cite{Gong_CVPR12}.The best (in bold blue), the second best (in blue).}
\label{tab:Unsuper_VGG}
\scalebox{0.85}{
\begin{tabular}{|r|c|c|c|c|c|c|c|c|c|c|c|c|}
\hline
      & A$\rightarrow$W   & A$\rightarrow$D   & A$\rightarrow$C   & W$\rightarrow$A   & W$\rightarrow$D   & W$\rightarrow$C   & D$\rightarrow$A   & D$\rightarrow$W   & D$\rightarrow$C   & C$\rightarrow$A   & C$\rightarrow$W   & C$\rightarrow$D   \\ \hline
1-NN-s & 60.9 & 52.3  & 70.1  & 62.4 & 83.9  &  57.5 & 57.0  & 86.7 & 48.0 & 81.9 & 65.9 & 55.6  \\
SVM-s     & 63.1 &  51.7 & 74.2  & 69.8  & 89.4  & 64.7 & 58.7 & 91.8 & 55.5 & 86.7 & 74.8 & 61.5   \\
GFK-PLS & 74.1 & 63.5 &  77.7 & 77.9  & \textbf{\color{blue}92.9}  & 71.3 & 69.9 & 92.4 & 64.0 & 86.2 & 76.5  &  66.5\\
SA    & {\color{Blue}76.0} & 64.9  & 77.1  &  76.6 & 90.4  & 70.7  & 69.0  & 90.5  &  62.3 & 83.9  & 76.0  & 66.2  \\
CORAL   & 74.8 & {\color{Blue}67.1}  & \textbf{\color{blue}79.0}  & {\color{Blue}81.2}  & {\color{Blue}92.6}  & {\color{Blue}75.2}  & {\color{Blue}75.8}  & \textbf{\color{blue}94.6}  & {\color{Blue}64.7}  & \textbf{\color{blue}89.4}  & {\color{Blue}77.6}  & {\color{Blue}67.6}   \\
\textbf{ILS} (1-NN) & \textbf{\color{blue}82.4} & \textbf{\color{blue}72.5}  & {\color{Blue}78.9}  & \textbf{\color{blue}85.9}  & 87.4  & \textbf{\color{blue}77.0}  & \textbf{\color{blue}79.2}  & {\color{Blue}94.2}  & \textbf{\color{blue}66.5}  & {\color{Blue}87.6}  & \textbf{\color{blue}84.4}  & \textbf{\color{blue}73.0} \\ \hline
\end{tabular}}
\vspace{-1mm}
\end{table*}

\begin{table*}[!t]
\centering
\small
\caption{Unsupervised domain adaptation results using VGG-FC7 features on Office$+$Caltech10~\cite{Gong_CVPR12} dataset with the evaluation setup of~\cite{Gong_CVPR12}.The best (in bold blue), the second best (in blue).}
\label{tab:Unsuper_FC7}
\scalebox{0.85}{
\begin{tabular}{|r|c|c|c|c|c|c|c|c|c|c|c|c|}
\hline
      & A$\rightarrow$W   & A$\rightarrow$D   & A$\rightarrow$C   & W$\rightarrow$A   & W$\rightarrow$D   & W$\rightarrow$C   & D$\rightarrow$A   & D$\rightarrow$W   & D$\rightarrow$C   & C$\rightarrow$A   & C$\rightarrow$W   & C$\rightarrow$D   \\ \hline
1-NN-s & 64.0 & 50.8 & 72.6 & 64.5 & 83.1 & 60.2 & 61.2 & 88.2 & 52.8 & 82.6 & 65.3 & 54.9 \\
SVM-s  & 68.0 & 51.8 & 76.2 & 70.1 & 87.4 & 65.5 & 58.7 & 91.2 & 56.0 & 86.7 & 74.8 & 61.3 \\
GFK-PLS & 74.0 & 57.6 & 76.6 & 75.0 & {\color{blue}89.6} & 62.1 & 67.5 & \textbf{{\color{blue}91.9}} & 62.9 & 84.1 & 73.6 & 63.4 \\
SA  & {\color{blue}75.0} & 60.7 & 76.2 & 74.6 & {88.8} & 67.5 & 66.0 & 89.5 & 59.4 & 82.6 & 73.6 & 63.2  \\
CORAL & 71.8 & {\color{blue}61.3} & \textbf{{\color{blue}78.6}} & {\color{blue}81.4} & \textbf{{\color{blue}90.1}} & {\color{blue}73.6} & {\color{blue}71.2} & 93.5 & {\color{blue}63.0} & {\textbf{\color{blue}88.6}} & {\color{blue}76.0} & {\color{blue}63.8}  \\
\textbf{ILS} (1-NN) & \textbf{{\color{blue}80.9}} & \textbf{{\color{blue}71.3}} & {\color{blue}78.4} & \textbf{{\color{blue}85.7}} & 84.8 & \textbf{{\color{blue}75.1}} & \textbf{{\color{blue}76.5}} & {\color{blue}91.8} & \textbf{{\color{blue}66.2}} & {\color{blue}87.1} & \textbf{{\color{blue}80.1}} & \textbf{{\color{blue}67.1}} \\ \hline
\end{tabular}}
\end{table*}

\noindent \textbf{Learned Transformations as Subspace Representations}: We consider both GFK~\citep{Gong_CVPR12} and SA~\citep{Basura_ICCV13} as DA-SL algorithms. Both these methods make use of PCA subspaces to adapt the domains. To the best of our knowledge,
there exists no through studies claiming that PCA is the method of choice to for GFK and SA. As a matter of fact, Gong \etal 
show that the performance of GFK can be improved if PLS\footnote{Despite using labeled data, this method falls under the unsupervised setting since it does not use the labeled target data.} algorithm is employed to define the source subspace~\citep{Gong_CVPR12}. 
Whether PCA or PLS is used to define the subspaces, identification of the subspaces is disjoint from  the domain adaptation technique
in GFK and SA. 
In contrast, transformations in the ILS algorithm are linked to the adaptation process. 
This makes a curious mind wondering whether the learned transformations in the ILS algorithm capture better structures for adaptation or not.
We empirically show that this is indeed the case by using the learned $\Mat{W}_s$ as the source subspace in GFK and SA.

\begin{figure}[t]
  \centering
  \includegraphics[width=0.7\textwidth]{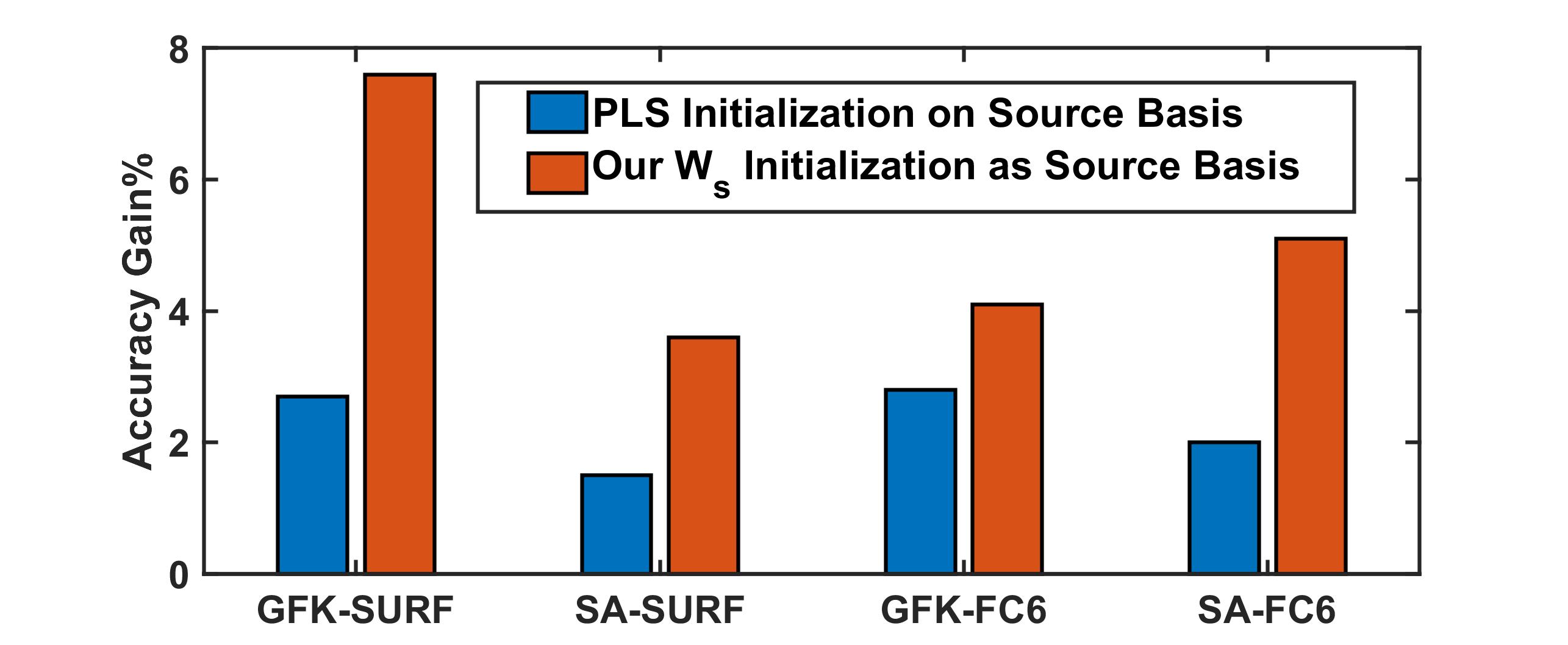}
 \caption{The accuracy gain on Office-Caltech dataset for GFK~\citep{Gong_CVPR12} and SA~\citep{Basura_ICCV13} when their initial PCA subspaces are replaced with PLS and our $W_s$ transformation matrices.}
 \label{fig:AccGain}
 \end{figure}

Figure~\ref{fig:AccGain} compares the accuracy gains over PCA spaces by using PLS and our $\Mat{W}_s$ initialization. It is clear that the highest classification accuracy gain is obtained by our $\Mat{W}_s$ initialization. This proves that $\Mat{W}_s$ is capable to learn a more favorable subspace representation for domain adaptation.

\subsubsection{PIE-Multiview Faces}

\begin{figure}[t!]
 \centering
 \scalebox{0.6}{
  \includegraphics[width=0.8\textwidth]{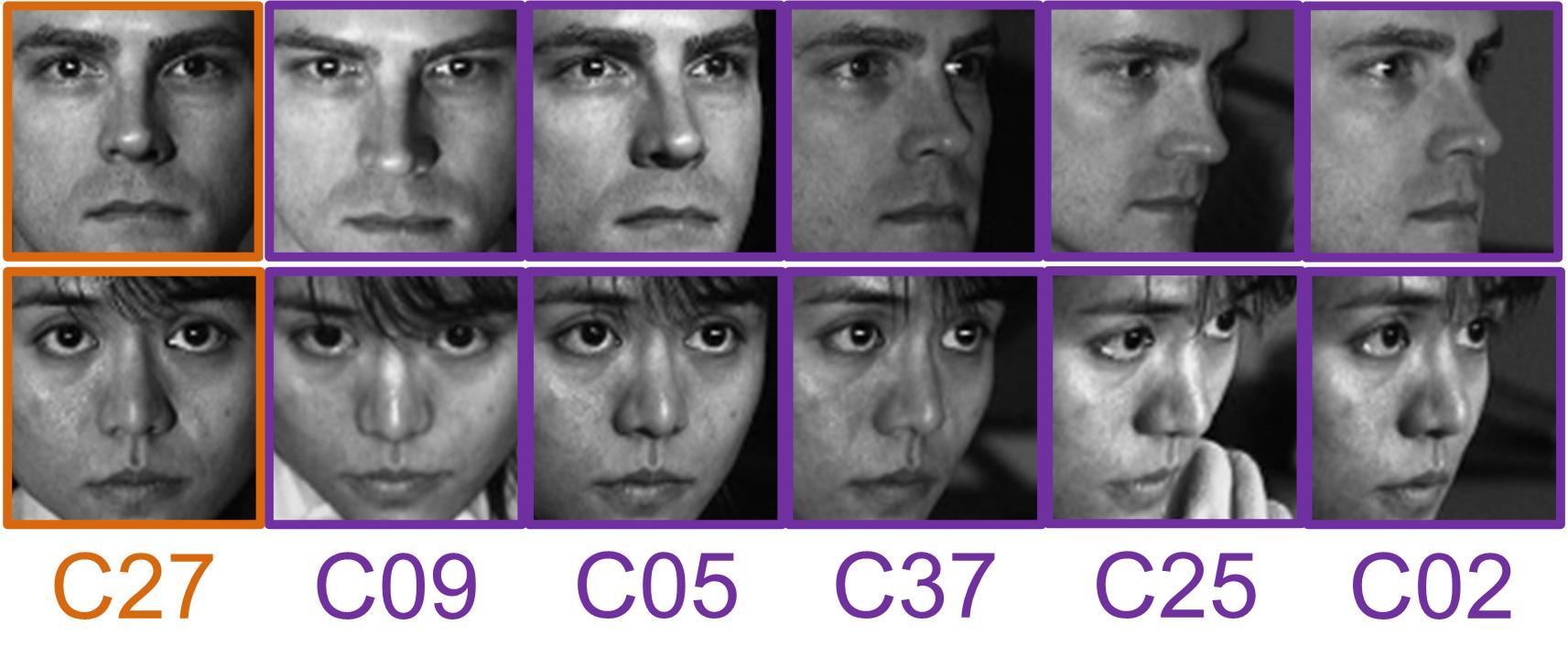}}
 \caption{Two instances of the PIE-Multiview face data. Here, the view from $C27$ is used as the source domain. Remaining views are considered to be the target for each transformation.}
 \label{fig:pie-faces}
 \end{figure}

The PIE Multiview dataset includes face images of 67 individuals captured from different views, illumination conditions, and expressions. 
In this experiment, we use the views $C27$ (looking forward) as the source domain and $C09$ (looking down), and the views $C05$, $C37$, $C02$, $C25$ (looking towards left in an increasing angle, see Fig.~\ref{fig:pie-faces}) as target domains. 
We expect the face inclination angle to reflect the complexity of transfer learning. We normalize the images to  32$\times$ 32 pixels and use the vectorized gray-scale images as features. 
Empirically, we observed that the GFK~\citep{Gong_CVPR12} and SA~\citep{Basura_ICCV13} reach better performances if the features are normalized
to have unit $\ell_2$ norm. We therefore use $\ell_2$ normalized features in our evaluations. 
The dimensionality of the subspaces for all the subspace based methods (\ie,~\cite{Gong_CVPR12,Basura_ICCV13}) including ours is 100. 

Table.~\ref{tab:PIE_Results} lists the classification accuracies with increasing angle of inclination. Our solution attains the  best scores for 4 views and the second best for the $C09$. With the increasing camera angle, the feature structure changes up to a certain extent. In other words, the features become heterogeneous. However, our algorithm boosts the accuracies 
even under such challenging conditions. 

\begin{table}[t!]
\centering
\small
\caption{PIE-Multiview results. The variation of performance w.r.t. face orientations when frontal face images are considered as the source domain.
}
\label{tab:PIE_Results}
\scalebox{0.85}{
\begin{tabular}{|r|c|c|c|c|c|}
\hline
\multicolumn{1}{|c|}{camera pose$\rightarrow$} & C09  & C05  & C37  & C25  & C02 \\ \hline
1-NN-s & 92.5 & 55.7 & 28.5 & 14.8 & 11.0 \\
SVM-s & 87.8 & 65.0 & 35.8 & 15.7 & {\color{Blue}16.7}  \\
GFK-PLS & 92.5 & 74.0 & 32.1 & 14.1 & 12.3 \\
SA  & \textbf{\color{blue}97.9} & {\color{Blue}85.9} & {\color{Blue}47.9} & {\color{Blue}16.6} & 13.9 \\
CORAL & 91.4 & 74.8 & 35.3 & 13.4 & 13.2  \\
\textbf{ILS} (1-NN) & {\color{Blue}96.6} & \textbf{\color{blue}88.3} & \textbf{\color{blue}72.9} & \textbf{\color{blue}28.4} & \textbf{\color{blue}34.8}   \\ \hline
\end{tabular}}
\end{table}

\section{Parameter Sensitivity and Orthogonality}

In all the above experiments, we keep $\lambda = 1$ (see Eq.~\ref{eqn:generalCost}). To analyze the sensitivity of our method to the changes in parameter $\lambda$, we performed an experiment using the unsupervised protocol. This is because the statistical loss plays a significant role in establishing the correspondence between the source and the target in the unsupervised DA. We consider two random splits from each of the Office+Caltech10 dataset along VGG-FC6 features here. 

\begin{figure}[!t]
 \centering
 \scalebox{0.6}{
  \includegraphics[width=1\textwidth]{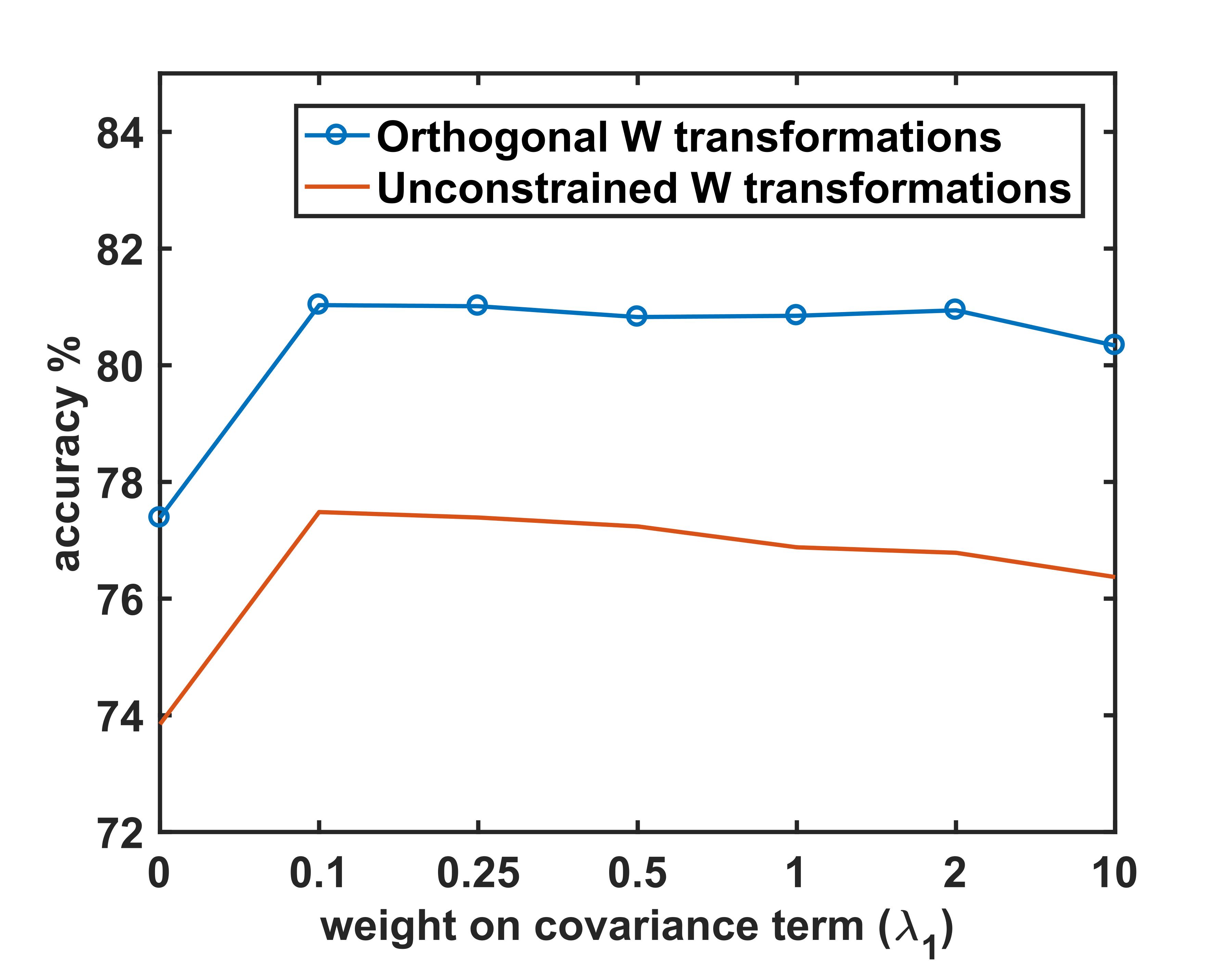}}
 \caption{Sensitivity to $\lambda$ in the unsupervised DA.}
 \label{fig:ParamSensitivity}
\end{figure}

Our results are shown in Fig.~\ref{fig:ParamSensitivity}. When $\lambda = 0$, no statistical loss term is considered. It is clear that for this case the performance drops considerably. For other values of $\lambda$, the performance is superior and there is little variation in performance. In other words, our method remains robust. 

We further investigate the benefit of orthogonality constraint on $\Mat{W}_s$ and $\Mat{W}_t$ against free-form and unconstrained  transformations. 
Using the orthogonality constraint provides a considerable performance gain as shown in Fig.~\ref{fig:ParamSensitivity}. 
While orthogonality makes the optimization more complicated, it seems it guides the learning to better uncovering the form of adaptation.

\section*{Conclusion}
In this paper, we proposed a solution for both semi-supervised and unsupervised Domain Adaptation (DA) problems. 
Our solution learns a latent space in which domain discrepancies are minimized. 
We showed that such a latent space can be obtained by  \textbf{1.} minimizing a notion of discriminatory power over the available labeled data 
while simultaneously \textbf{2.} matching statistical properties across the domains. 
To determine the latent space, we modeled the learning problem as a minimization problem on Riemannian manifolds 
and solved it using optimization techniques on matrix manifolds.

Empirically, we showed that the proposed method outperformed state-of-the-art DA solutions in semi-supervised and unsupervised settings. With the proposed framework we see possibilities of extending our solution to large scale datasets with stochastic optimization techniques, multiple source DA and for domain generalization~\citep{Ghifary_TPAMI16,Gan_CVPR16}. In terms of algorithmic extensions we look forward to use dictionary learning~\citep{Koniusz_SparseCode16} and higher order statistics matching.


\bibliography{references}

\begin{thebibliography}{55}
\providecommand{\natexlab}[1]{#1}
\providecommand{\url}[1]{\texttt{#1}}
\expandafter\ifx\csname urlstyle\endcsname\relax
  \providecommand{\doi}[1]{doi: #1}\else
  \providecommand{\doi}{doi: \begingroup \urlstyle{rm}\Url}\fi

\bibitem[Absil et~al.(2009)Absil, Mahony, and Sepulchre]{Absil_Book}
P-A Absil, Robert Mahony, and Rodolphe Sepulchre.
\newblock \emph{Optimization algorithms on matrix manifolds}.
\newblock Princeton University Press, 2009.

\bibitem[Baktashmotlagh et~al.(2016)Baktashmotlagh, Harandi, and
  Salzmann]{Masha_ICCV14}
Mahsa Baktashmotlagh, Mehrtash Harandi, and Mathieu Salzmann.
\newblock Distribution-matching embedding for visual domain adaptation.
\newblock \emph{Journal of Machine Learning Research}, 17\penalty0
  (108):\penalty0 1--30, 2016.

\bibitem[Bay et~al.(2006)Bay, Tuytelaars, and Van~Gool]{Bay_SURF06}
Herbert Bay, Tinne Tuytelaars, and Luc Van~Gool.
\newblock Surf: Speeded up robust features.
\newblock In \emph{European conference on computer vision}, pages 404--417.
  Springer, 2006.

\bibitem[Bonnabel(2013)]{Bonnabel_Stoch13}
S.~Bonnabel.
\newblock Stochastic gradient descent on {R}iemannian manifolds.
\newblock \emph{IEEE Transactions on Automatic Control}, 58\penalty0
  (9):\penalty0 2217--2229, 2013.

\bibitem[Borgwardt et~al.(2006)Borgwardt, Gretton, Rasch, Kriegel, Schoelkopf,
  and Smola]{Borgwardt_Bio06}
Karsten Borgwardt, Arthur Gretton, Malte~J. Rasch, Hans-Peter Kriegel, Bernhard
  Schoelkopf, and Alexander Smola.
\newblock Integrating structured biological data by kernel maximum mean
  discrepancy.
\newblock \emph{Bioinformatics}, 22:\penalty0 49--57, 2006.

\bibitem[Boumal et~al.(2014)Boumal, Mishra, Absil, and Sepulchre]{manopt}
N.~Boumal, B.~Mishra, P.-A. Absil, and R.~Sepulchre.
\newblock {M}anopt, a {M}atlab toolbox for optimization on manifolds.
\newblock \emph{Journal of Machine Learning Research}, 15:\penalty0 1455--1459,
  2014.
\newblock URL \url{http://www.manopt.org}.

\bibitem[Boyd and Vandenberghe(2004)]{Boyd_Convex04}
Stephen Boyd and Lieven Vandenberghe.
\newblock \emph{Convex Optimization}.
\newblock Cambridge University Press, New York, NY, USA, 2004.

\bibitem[Brookes(2005)]{MatrixRefManual}
Mike Brookes.
\newblock The matrix reference manual.
\newblock \emph{Imperial College London}, 2005.

\bibitem[Chen et~al.(2015)Chen, Huang, Feris, Brown, Dong, and
  Yan]{Qiang_CVPR15}
Qiang Chen, Junshi Huang, Rogerio Feris, Lisa~M Brown, Jian Dong, and Shuicheng
  Yan.
\newblock Deep domain adaptation for describing people based on fine-grained
  clothing attributes.
\newblock In \emph{Proc. IEEE Conference on Computer Vision and Pattern
  Recognition (CVPR)}, pages 5315--5324, 2015.

\bibitem[Cherian et~al.(2013)Cherian, Sra, Banerjee, and
  Papanikolopoulos]{Anoop_TPAMI13}
A.~Cherian, S.~Sra, A.~Banerjee, and N.~Papanikolopoulos.
\newblock Jensen-bregman logdet divergence with application to efficient
  similarity search for covariance matrices.
\newblock \emph{IEEE Transactions on Pattern Analysis and Machine
  Intelligence}, 35\penalty0 (9):\penalty0 2161--2174, Sept 2013.

\bibitem[Cherian and Sra(2016)]{Cherrian_SPD2016}
Anoop Cherian and Suvrit Sra.
\newblock Positive definite matrices: data representation and applications to
  computer vision.
\newblock \emph{Algorithmic Advances in Riemannian Geometry and Applications:
  For Machine Learning, Computer Vision, Statistics, and Optimization},
  page~93, 2016.

\bibitem[Cherian et~al.(2016)Cherian, Morellas, and
  Papanikolopoulos]{Cherrian_LogDet16}
Anoop Cherian, Vassilios Morellas, and Nikolaos Papanikolopoulos.
\newblock Bayesian nonparametric clustering for positive definite matrices.
\newblock \emph{IEEE transactions on pattern analysis and machine
  intelligence}, 38\penalty0 (5):\penalty0 862--874, 2016.

\bibitem[Daum{\'e}~III et~al.(2010)Daum{\'e}~III, Kumar, and Saha]{Daume_DA10}
Hal Daum{\'e}~III, Abhishek Kumar, and Avishek Saha.
\newblock Frustratingly easy semi-supervised domain adaptation.
\newblock In \emph{Proceedings of the 2010 Workshop on Domain Adaptation for
  Natural Language Processing}, pages 53--59, 2010.

\bibitem[Davis et~al.(2007)Davis, Kulis, Jain, Sra, and Dhillon]{Davis_ICML07}
Jason~V. Davis, Brian Kulis, Prateek Jain, Suvrit Sra, and Inderjit~S. Dhillon.
\newblock Information-theoretic metric learning.
\newblock In \emph{ICML}, pages 209--216, Corvalis, Oregon, USA, June 2007.

\bibitem[Duan et~al.(2012)Duan, Xu, and Tsang]{Duan_ICML12}
Lixin Duan, Dong Xu, and Ivor~W. Tsang.
\newblock Learning with augmented features for heterogeneous domain adaptation.
\newblock In \emph{Proc. Int. Conference on Machine Learning (ICML)}, pages
  711--718, June 2012.

\bibitem[Edelman et~al.(1998)Edelman, Arias, and Smith]{Edelman_1998}
Alan Edelman, Tom{\'a}s~A Arias, and Steven~T Smith.
\newblock The geometry of algorithms with orthogonality constraints.
\newblock \emph{SIAM journal on Matrix Analysis and Applications}, 20\penalty0
  (2):\penalty0 303--353, 1998.

\bibitem[Fernando et~al.(2013)Fernando, Habrard, Sebban, and
  Tuytelaars]{Basura_ICCV13}
B.~Fernando, A.~Habrard, M.~Sebban, and T.~Tuytelaars.
\newblock Unsupervised visual domain adaptation using subspace alignment.
\newblock In \emph{Proc. Int. Conference on Computer Vision (ICCV)}, pages
  2960--2967, 2013.

\bibitem[Fernando et~al.(2015)Fernando, Tommasi, and
  Tuytelaars]{Fernando_PRL15}
Basura Fernando, Tatiana Tommasi, and Tinne Tuytelaars.
\newblock Joint cross-domain classification and subspace learning for
  unsupervised adaptation.
\newblock \emph{Pattern Recognition Letters}, 65:\penalty0 60 -- 66, 2015.

\bibitem[Gan et~al.(2016)Gan, Yang, and Gong]{Gan_CVPR16}
Chuang Gan, Tianbao Yang, and Boqing Gong.
\newblock Learning attributes equals multi-source domain generalization.
\newblock In \emph{Proc. IEEE Conference on Computer Vision and Pattern
  Recognition (CVPR)}, pages 87--97, 2016.

\bibitem[Ganin and Lempitsky(2015)]{Ganin_ICML15}
Yaroslav Ganin and Victor Lempitsky.
\newblock Unsupervised domain adaptation by backpropagation.
\newblock In \emph{Proc. Int. Conference on Machine Learning (ICML)}, pages
  1180--1189, 2015.

\bibitem[Ghifary et~al.(2016)Ghifary, Balduzzi, Kleijn, and
  Zhang]{Ghifary_TPAMI16}
M.~Ghifary, D.~Balduzzi, W.~B. Kleijn, and M.~Zhang.
\newblock Scatter component analysis: A unified framework for domain adaptation
  and domain generalization.
\newblock \emph{IEEE Trans. Pattern Analysis and Machine Intelligence},
  PP\penalty0 (99):\penalty0 1--1, 2016.

\bibitem[Gong et~al.(2012)Gong, Shi, Sha, and Grauman]{Gong_CVPR12}
B.~Gong, Y.~Shi, F.~Sha, and K.~Grauman.
\newblock Geodesic flow kernel for unsupervised domain adaptation.
\newblock In \emph{Proc. IEEE Conference on Computer Vision and Pattern
  Recognition (CVPR)}, pages 2066--2073, 2012.

\bibitem[Gopalan et~al.(2011)Gopalan, Li, and Chellappa]{Gopalan_ICCV11}
R.~Gopalan, Ruonan Li, and R.~Chellappa.
\newblock Domain adaptation for object recognition: An unsupervised approach.
\newblock In \emph{Proc. Int. Conference on Computer Vision (ICCV)}, pages
  999--1006, 2011.

\bibitem[Guillemin and Pollack(2010)]{Guillemin_Book}
Victor Guillemin and Alan Pollack.
\newblock \emph{Differential topology}, volume 370.
\newblock American Mathematical Soc., 2010.

\bibitem[Harandi and Fernando(2016)]{Harandi_GB16}
Mehrtash Harandi and Basura Fernando.
\newblock Generalized backpropagation, {\'{e}}tude de cas: Orthogonality.
\newblock \emph{CoRR}, abs/1611.05927, 2016.

\bibitem[Harandi et~al.(2016)Harandi, Salzmann, and Hartley]{Harandi_SPD16}
Mehrtash~Tafazzoli Harandi, Mathieu Salzmann, and Richard~I. Hartley.
\newblock Dimensionality reduction on {SPD} manifolds: The emergence of
  geometry-aware methods.
\newblock \emph{CoRR}, abs/1605.06182, 2016.

\bibitem[He et~al.(2016)He, Zhang, Ren, and Sun]{He_CVPR16}
Kaiming He, Xiangyu Zhang, Shaoqing Ren, and Jian Sun.
\newblock Deep residual learning for image recognition.
\newblock In \emph{Proc. IEEE Conference on Computer Vision and Pattern
  Recognition (CVPR)}, pages 770--778, June 2016.

\bibitem[Herath et~al.(2017)Herath, Harandi, and Porikli]{Herath_Survey17}
Samitha Herath, Mehrtash Harandi, and Fatih Porikli.
\newblock Going deeper into action recognition: A survey.
\newblock \emph{Image and Vision Computing}, 60:\penalty0 4 -- 21, 2017.
\newblock Regularization Techniques for High-Dimensional Data Analysis.

\bibitem[Hoffman et~al.(2013)Hoffman, Rodner, Donahue, Saenko, and
  Darrell]{Hoffman_ICLR2013}
Judy Hoffman, Erik Rodner, Jeff Donahue, Kate Saenko, and Trevor Darrell.
\newblock Efficient learning of domain-invariant image representations.
\newblock In \emph{International Conference on Learning Representations}, 2013.

\bibitem[Hoffman et~al.(2014)Hoffman, Rodner, Donahue, Kulis, and
  Saenko]{Hoffman_IJCV14}
Judy Hoffman, Erik Rodner, Jeff Donahue, Brian Kulis, and Kate Saenko.
\newblock Asymmetric and category invariant feature transformations for domain
  adaptation.
\newblock \emph{Int. Journal of Computer Vision}, 109\penalty0 (1):\penalty0
  28--41, 2014.

\bibitem[Hubert~Tsai et~al.(2016)Hubert~Tsai, Yeh, and Frank~Wang]{Tsai_CVPR16}
Yao-Hung Hubert~Tsai, Yi-Ren Yeh, and Yu-Chiang Frank~Wang.
\newblock Learning cross-domain landmarks for heterogeneous domain adaptation.
\newblock In \emph{The IEEE Conference on Computer Vision and Pattern
  Recognition (CVPR)}, pages 5081--5090, June 2016.

\bibitem[James(1976)]{James_Stiefel76}
Ioan~Mackenzie James.
\newblock \emph{The topology of Stiefel manifolds}, volume~24.
\newblock Cambridge University Press, 1976.

\bibitem[Koniusz and Cherian(2016)]{Koniusz_SparseCode16}
Piotr Koniusz and Anoop Cherian.
\newblock Sparse coding for third-order super-symmetric tensor descriptors with
  application to texture recognition.
\newblock \emph{Proc. IEEE Conference on Computer Vision and Pattern
  Recognition (CVPR)}, page 5395, 2016.

\bibitem[Koniusz et~al.(2017)Koniusz, Tas, and Porikli]{Koniusz_CVPR16}
Piotr Koniusz, Yusuf Tas, and Fatih Porikli.
\newblock Domain adaptation by mixture of alignments of second-or higher-order
  scatter tensors.
\newblock \emph{Proc. IEEE Conference on Computer Vision and Pattern
  Recognition (CVPR)}, 2017.

\bibitem[K{\"o}stinger et~al.(2012)K{\"o}stinger, Hirzer, Wohlhart, Roth, and
  Bischof]{Koestinger_CVPR12}
Martin K{\"o}stinger, Martin Hirzer, Paul Wohlhart, Peter~M Roth, and Horst
  Bischof.
\newblock Large scale metric learning from equivalence constraints.
\newblock In \emph{Computer Vision and Pattern Recognition (CVPR), 2012 IEEE
  Conference on}, pages 2288--2295, 2012.

\bibitem[Krishnan et~al.(2011)Krishnan, Williams, McIntosh, and
  Abdi]{Krishnan_PLS11}
Anjali Krishnan, Lynne~J Williams, Anthony~Randal McIntosh, and Herv{\'e} Abdi.
\newblock Partial least squares (pls) methods for neuroimaging: a tutorial and
  review.
\newblock \emph{Neuroimage}, 56\penalty0 (2):\penalty0 455--475, 2011.

\bibitem[Krizhevsky et~al.(2012)Krizhevsky, Sutskever, and
  Hinton]{Krizhevsky_NIPS12}
Alex Krizhevsky, Ilya Sutskever, and Geoffrey~E. Hinton.
\newblock Imagenet classification with deep convolutional neural networks.
\newblock In \emph{Proc. Advances in Neural Information Processing Systems
  (NIPS)}, pages 1097--1105, 2012.

\bibitem[Kulis et~al.(2011)Kulis, Saenko, and Darrell]{Kulis_CVPR11}
B.~Kulis, K.~Saenko, and T.~Darrell.
\newblock What you saw is not what you get: Domain adaptation using asymmetric
  kernel transforms.
\newblock In \emph{Proc. IEEE Conference on Computer Vision and Pattern
  Recognition (CVPR)}, pages 1785--1792, June 2011.

\bibitem[Lee(2003)]{Lee_Book}
John~M Lee.
\newblock Smooth manifolds.
\newblock In \emph{Introduction to Smooth Manifolds}, pages 1--29. Springer,
  2003.

\bibitem[Long et~al.(2016)Long, Wang, Cao, Sun, and Yu]{Long_Deep16}
M.~Long, J.~Wang, Y.~Cao, J.~Sun, and P.~S. Yu.
\newblock Deep learning of transferable representation for scalable domain
  adaptation.
\newblock \emph{IEEE Transactions on Knowledge and Data Engineering},
  28\penalty0 (8):\penalty0 2027--2040, Aug 2016.

\bibitem[Ni et~al.(2013)Ni, Qiu, and Chellappa]{Ni_Subspace13}
Jie Ni, Qiang Qiu, and Rama Chellappa.
\newblock Subspace interpolation via dictionary learning for unsupervised
  domain adaptation.
\newblock In \emph{Proceedings of the IEEE Conference on Computer Vision and
  Pattern Recognition}, pages 692--699, 2013.

\bibitem[Oh~Song et~al.(2016)Oh~Song, Xiang, Jegelka, and
  Savarese]{Song_CVPR16}
Hyun Oh~Song, Yu~Xiang, Stefanie Jegelka, and Silvio Savarese.
\newblock Deep metric learning via lifted structured feature embedding.
\newblock In \emph{Proc. IEEE Conference on Computer Vision and Pattern
  Recognition (CVPR)}, June 2016.

\bibitem[Pan et~al.(2011)Pan, Tsang, Kwok, and Yang]{Pan_TNN11}
S.~J. Pan, I.~W. Tsang, J.~T. Kwok, and Q.~Yang.
\newblock Domain adaptation via transfer component analysis.
\newblock \emph{IEEE Transactions on Neural Networks}, 22\penalty0
  (2):\penalty0 199--210, 2011.

\bibitem[Patel et~al.(2015)Patel, Gopalan, Li, and Chellappa]{Patel_Survey15}
V.~M. Patel, R.~Gopalan, R.~Li, and R.~Chellappa.
\newblock Visual domain adaptation: A survey of recent advances.
\newblock \emph{IEEE Signal Processing Magazine}, 32\penalty0 (3):\penalty0
  53--69, 2015.

\bibitem[Pennec et~al.(2006)Pennec, Fillard, and Ayache]{Pennec_IJCV06}
Xavier Pennec, Pierre Fillard, and Nicholas Ayache.
\newblock A riemannian framework for tensor computing.
\newblock \emph{Int. Journal of Computer Vision}, 66\penalty0 (1):\penalty0
  41--66, 2006.

\bibitem[Russakovsky et~al.(2015)Russakovsky, Deng, Su, Krause, Satheesh, Ma,
  Huang, Karpathy, Khosla, Bernstein, Berg, and Fei-Fei]{ILSVRC15}
Olga Russakovsky, Jia Deng, Hao Su, Jonathan Krause, Sanjeev Satheesh, Sean Ma,
  Zhiheng Huang, Andrej Karpathy, Aditya Khosla, Michael Bernstein,
  Alexander~C. Berg, and Li~Fei-Fei.
\newblock {ImageNet Large Scale Visual Recognition Challenge}.
\newblock \emph{Int. Journal of Computer Vision}, 115\penalty0 (3):\penalty0
  211--252, 2015.

\bibitem[Sa et~al.(2015)Sa, Re, and Olukotun]{Sa_ICML15}
Christopher~D Sa, Christopher Re, and Kunle Olukotun.
\newblock Global convergence of stochastic gradient descent for some non-convex
  matrix problems.
\newblock In \emph{Proc. Int. Conference on Machine Learning (ICML)}, pages
  2332--2341, 2015.

\bibitem[Saenko et~al.(2010)Saenko, Kulis, Fritz, and Darrell]{Saenko_ECCV10}
Kate Saenko, Brian Kulis, Mario Fritz, and Trevor Darrell.
\newblock Adapting visual category models to new domains.
\newblock In \emph{Proc. European Conference on Computer Vision (ECCV)}, pages
  213--226, 2010.

\bibitem[Shi et~al.(2010)Shi, Liu, Fan, Philip, and Zhu]{Shi_ICDM10}
Xiaoxiao Shi, Qi~Liu, Wei Fan, S~Yu Philip, and Ruixin Zhu.
\newblock Transfer learning on heterogenous feature spaces via spectral
  transformation.
\newblock In \emph{2010 IEEE international conference on data mining}, pages
  1049--1054, 2010.

\bibitem[Shimodaira(2000)]{Shimodaira_JSPI00}
Hidetoshi Shimodaira.
\newblock Improving predictive inference under covariate shift by weighting the
  log-likelihood function.
\newblock \emph{Journal of Statistical Planning and Inference}, 90\penalty0
  (2):\penalty0 227 -- 244, 2000.

\bibitem[Simonyan and Zisserman(2014)]{Simonyan_VGG14}
Karen Simonyan and Andrew Zisserman.
\newblock Very deep convolutional networks for large-scale image recognition.
\newblock \emph{arXiv preprint arXiv:1409.1556}, 2014.

\bibitem[Sun et~al.(2016)Sun, Feng, and Saenko]{SUN_Return15}
Baochen Sun, Jiashi Feng, and Kate Saenko.
\newblock Return of frustratingly easy domain adaptation.
\newblock In \emph{Thirtieth AAAI Conference on Artificial Intelligence}, 2016.

\bibitem[Torralba and Efros(2011)]{Torralba_CVPR11}
A.~Torralba and A.~A. Efros.
\newblock Unbiased look at dataset bias.
\newblock In \emph{Proc. IEEE Conference on Computer Vision and Pattern
  Recognition (CVPR)}, pages 1521--1528, 2011.

\bibitem[Tzeng et~al.(2014)Tzeng, Hoffman, Zhang, Saenko, and
  Darrell]{Tzeng_Deep14}
Eric Tzeng, Judy Hoffman, Ning Zhang, Kate Saenko, and Trevor Darrell.
\newblock Deep domain confusion: Maximizing for domain invariance.
\newblock \emph{arXiv preprint arXiv:1412.3474}, 2014.

\bibitem[Wang and Mahadevan(2011)]{Wang_IJCAI11}
Chang Wang and Sridhar Mahadevan.
\newblock Heterogeneous domain adaptation using manifold alignment.
\newblock In \emph{Proceedings of the Twenty-Second International Joint
  Conference on Artificial Intelligence - Volume Volume Two}, IJCAI'11, pages
  1541--1546, 2011.

\end{thebibliography}

\clearpage

\section*{Appendix}

In this appendix, we provide more details on the product geometry of the problem discussed in \textsection~\ref{sec:proposed_method} and also
the form of gradients required to perform Riemannian optimization.  

\subsection*{A. Product Topology}

As the constraints of the optimization problem depicted in Eq.~\ref{eqn:generalCost} are indeed Riemannian manifolds, the whole set of 
constraints can be given a Riemannian structure through the concept of product topology. In particular, the 
constraints can be modeled as
 
\begin{equation}
\label{eqn:prodTopology_ap}
\mathcal{M}_{prod.} = \ST{s}{p} \times \ST{t}{p} \times \SPD{p} \times \mathbb{R}^{N_p},
\end{equation} 

\noindent The tangent space of such a product topology~\citep{Lee_Book,Guillemin_Book} could be written as,  

\begin{equation}
\label{eqn:tangentProd_ap}
\mathcal T_{(\Mat{W}_s,\Mat{W}_t,\Mat{M},\Vec{\epsilon})} \mathcal{M}_{prod.} = T_{\Mat{W}_s}\ST{s}{p} \times T_{\Mat{W}_t}\ST{t}{p} \times T_{\Mat{M}}\SPD{p} \times T_{\Vec{\epsilon}}\mathbb{R}^{N_p}.
\end{equation} 

\noindent In Table~\ref{tab:ToolsProduct}, the metric and, the form of Riemannian gradient and the retraction for $\mathcal{M}_{prod.}$
are provided. 

\subsection*{B. Derivations}

We recall that the cost function depicted in Eq.~\ref{eqn:generalCost} consists of two parts, namely 
$\mathcal{L}_{d}$ and $\mathcal{L}_{u}$.
Here, $\mathcal{L}_{d}$ is a measure of dissimilarity between labeled samples. The term
$\mathcal{L}_{u}$ quantifies a notion of  statistical difference between the source and target samples in the latent space. 
We provide the gradients of Eq.~\ref{eqn:generalCost} with respect to its parameters below. This, as discussed in \textsection~\ref{sec:opt}
is required for Riemannian optimization. 

\subsection*{Derivative of soft-margin $\ell_\beta$}

\noindent We recall that $\mathcal{L}_d$ has the following form,

\begin{align}
\label{eqn:soft_margin_cost_ap}
\mathcal{L}_{d} &= \frac{1}{N_p} \sum\limits_{k=1}^{N_p} 
\ell_\beta\big(\Mat{M},y_k,\Vec{z}_{1,k} - \Vec{z}_{2,k},1+y_k\epsilon_k \big)  + r(\Mat{M}) + \frac{1}{N_p} \sqrt{\sum \epsilon_k^2}\;,
\end{align}
with 

\begin{align}
\label{eqn:ell_b_ap}
\hspace{-2ex} \ell_\beta\big(\Mat{M},y,\Vec{x},u \big)  = \frac{1}{\beta}\log\Big(1+\exp\big(\beta y(\Vec{x}^T\Mat{M}\Vec{x} -u )\big)\Big).
\end{align}

\noindent In Eq.~\ref{eqn:soft_margin_cost_ap}, $y_k$ denotes whether the $k$-th pair is similar or dissimilar (\ie, 
$y_k=+1$ if $\Vec{z}_{1,k}$ and $\Vec{z}_{2,k}$ are from the same class and $y_k=-1$ otherwise).

For the sake of discussion, assume $\Vec{z}_{1,k}$ and $\Vec{z}_{2,k}$ are embedded from the source and target domains, respectively. 
That is  $\Vec{z}_{1,k} = \Mat{W}_s^T\Vec{x}^s_i$ and $\Vec{z}_{2,k} = \Mat{W}_t^T\Vec{x}^t_j$. 
By expanding $\ell_\beta$ for such a pair, we get

\begin{align}
\label{eqn:ell_beta_ap}
\ell_\beta(\Mat{M},y_k,\Vec{z}_{1,k}-\Vec{z}_{2,k},1+y_k\epsilon_k) 
&= \frac{1}{\beta}\log(1+\exp(\beta y_k((\Vec{z}_{1,k}-\Vec{z}_{2,k})^T\Mat{M}(\Vec{z}_{1,k}-\Vec{z}_{2,k})-1-y_k\epsilon_k)))
\end{align}

\noindent
To simplify the presentation, we define $d(\Mat{M},\Mat{W}_s,\Mat{W_t}) = (\Vec{z}_{1,k}-\Vec{z}_{2,k})^T\Mat{M}(\Vec{z}_{1,k}-\Vec{z}_{2,k})$ and $r = \exp(\beta y_k((\Vec{z}_{1,k}-\Vec{z}_{2,k})^T\Mat{M}(\Vec{z}_{1,k}-\Vec{z}_{2,k})-y_k\epsilon_k-1))$. We provide the gradients of Eq.~\ref{eqn:ell_beta_ap} with respect to $\Mat{M}$, $\Mat{W_t}$, $\Mat{W_s}$ and the slack $\epsilon_k$ below. 

\subsubsection*{Derivative  w.r.t. $\Mat{M}$}

\begin{align}
\label{eqn:paramEllBeta}
\nabla_{\Mat{M}}\ell_\beta &= \frac{y_k r}{(1+r)} \nabla_{\Mat{M}}d(\Mat{M}) \nonumber\\
&= \frac{y_k r}{(1+ r)} (\Mat{W}_s^T{\Vec{x}^s_i}-\Mat{W}_t^T\Vec{x}^t_j)({\Vec{x}^s_i}^T\Mat{W}_s - {\Vec{x}^t_j}^T\Mat{W}_t) \nonumber\\
&= y_k(1+r^{-1})^{-1}(\Mat{W}_s^T{\Vec{x}^s_i}-\Mat{W}_t^T\Vec{x}^t_j)({\Vec{x}^s_i}^T\Mat{W}_s - {\Vec{x}^t_j}^T\Mat{W}_t).
\end{align}

\begin{table*}[!t]
\caption{Riemannian metric, gradient and retraction on the proposed Product Manifold in Eq.~\ref{eqn:prodTopology_ap}. The Riemannian metrics $g_{\nu_s}$ and $g_{\nu_t}$ are respectively defined on the Stiefel manifolds of $\Mat{W}_s$ and $\Mat{W}_t$. Furthermore, the Riemannian metrics $g_S$ and $g_E$ are respectively on the SPD manifold and the Euclidean manifold. As we have used in the main text,  $\xi$ and $\varsigma$ are elements from the tangent spaces of the corresponding manifolds. Here, 
$\mathrm{uf}(\Mat{A}) = \Mat{A}(\Mat{A}^T\Mat{A})^{-1/2}$, which yields an orthogonal matrix, $\mathrm{sym}(\Mat{A}) = \frac{1}{2}(\Mat{A} + \Mat{A}^T)$ and
$\expm(\cdot)$ denotes the matrix exponential.
}
\label{tab:ToolsProduct}
\centering
\scalebox{0.7}{
\begin{tabular}{|c|c|}
\hline
{\bf } &{\bf $\mathcal{M}_{prod.}$} \\
\hline 
Matrix representation 					&\big($\Mat{W}_s , \Mat{W}_t , \Mat{M} , \Vec{\epsilon} $\big)	\\
Riemannian metric						&$g_{\nu_s}(\varsigma_s,\xi_s) + g_{\nu_t}(\varsigma_t,\xi_t) + g_{\mathcal{S}}(\varsigma_M,\xi_M) + g_E(\varsigma_E,\xi_E)$\\
Riemannian gradient						&\Big($\nabla_{\Mat{W}_s}(f)  - \Mat{W}_s\mathrm{sym}\Big(\Mat{W}_s^T \nabla_{\Mat{W}_s}(f)\Big)$, $\nabla_{\Mat{W}_t}(f)  - \Mat{W}_t\mathrm{sym}\Big(\Mat{W}_t^T \nabla_{\Mat{W}_t}(f)\Big)$, $\Mat{M}\mathrm{sym}\Big(\nabla_\Mat{M}(f)\Big)\Mat{M}$, $\nabla_{\Vec{\epsilon}}(f)$\Big)                                              \\
Retraction								&\big($\mathrm{uf}(\Mat{W}_s + \xi_s)$, $\mathrm{uf}(\Mat{W}_t + \xi_t)$, $\Mat{M}^\frac{1}{2}\expm(\Mat{M}^{-\frac{1}{2}}\xi_M\Mat{M}^{-\frac{1}{2}})\Mat{M}^\frac{1}{2}$, $\mathbf{I}_p$\big)\\
\hline
\end{tabular}}
\label{tab:riemannian_tools}
\end{table*}

\subsubsection*{Derivative w.r.t. $\Mat{W}_s$ (or w.r.t. $\Mat{W}_t$)} 

\begin{align}
\label{eqn:paramWs}
\nabla_{\Mat{W}_s}\ell_\beta &= \frac{y_k r}{(1+ r)} \nabla_{\Mat{W}_s}d(\Mat{W}_s) \\
&= 2\frac{y_k r}{(1+ r)} \Vec{x}_i^s ({\Vec{x}^s_i}^T\Mat{W}_s - {\Vec{x}^t_j}^T\Mat{W}_t)\Mat{M} \nonumber \\
&= 2y_k(1+r^{-1})^{-1}\Vec{x}_i^s ({\Vec{x}^s_i}^T\Mat{W}_s - {\Vec{x}^t_j}^T\Mat{W}_t)\Mat{M}.
\end{align}

\noindent For the case where both the pair instances are from the same domain (\ie  $\Vec{z}_{1,k} = \Mat{W}_s^T\Vec{x}^s_i$ and $\Vec{z}_{2,k} = \Mat{W}_s^T\Vec{x}^s_j$), it could be shown that,

\begin{align}
\label{eqn:teqs}
\nabla_{\Mat{W}_s}d(\Mat{W}_s)  = 2y_k(\Vec{x}_i^s - \Vec{x}_j^s)({\Vec{x}^s_i}^T- {\Vec{x}^s_j}^T)\Mat{W}_s\Mat{M}.
\end{align}

\noindent Considering Eq.~\ref{eqn:paramWs} and Eq.~\ref{eqn:teqs},

\begin{align}
\label{eqn:paramWs2}
\nabla_{\Mat{W}_s}\ell_\beta &= 2\frac{y_k r}{(1+ r)} (\Vec{x}_i^s - \Vec{x}_j^s)({\Vec{x}^s_i}^T- {\Vec{x}^s_j}^T)\Mat{W}_s\Mat{M}. \nonumber\\
&= 2y_k(1+r^{-1})^{-1} (\Vec{x}_i^s - \Vec{x}_j^s)({\Vec{x}^s_i}^T- {\Vec{x}^s_j}^T)\Mat{W}_s\Mat{M}.
\end{align}

\subsubsection*{Derivative w.r.t. a Slack variable $\epsilon_k$.}

\smallskip
\noindent The slacks by origin are non-negative. To avoid using a non-negative constraint we make the substitution $\epsilon_k = e^{v_k}$ to Eq.~\ref{eqn:ell_beta_ap}.

\begin{align}
\label{eqn:slackell}
\therefore \ell_\beta = \frac{1}{\beta}\log(1+\exp(\beta y_k(d(\Mat{M},\Mat{W}_s,\Mat{W_t},v_k)-1-y_k e^{v_k})))
\end{align}

\smallskip 
\noindent The derivative of Eq.~\ref{eqn:slackell} w.r.t. $v_k$,

\begin{align}
\nabla_{{v}_k} \ell_\beta &= \frac{-e^{v_k}\exp(\beta y_k(d(\Mat{M},\Mat{W}_s,\Mat{W_t},v_k)-1-y_k e^{v_k}))}{(1+\exp(\beta y_k(d(\Mat{M},\Mat{W}_s,\Mat{W_t},v_k)-1- y_k e^{v_k}))} \nonumber \\
&= \frac{-e^{v_k}r}{(1+r)} = -e^{v_k}(1+r^{-1})^{-1}
\end{align}

\begin{table}[!t]
\centering
\caption{Summarized Gradients. Note we use $r = \exp(\beta y_k(d(\Mat{M})-y_k\epsilon_k-1))$. }
\label{tab:SummaryGrad_ap}
\begin{tabular}{|c|l|}
\hline
$\nabla_{\Mat{W}_s}\ell_\beta$ ; $x_i^s \in \mathbb{R}^s $, $x_j^t \in \mathbb{R}^t$  & 
$2y_k(1 + r^{-1})^{-1}\Vec{x}^s_i({\Vec{x}^s_i}^T\Mat{W}_s - {\Vec{x}^t_j}^T\Mat{W}_t)\Mat{M} 
$ \\ \hline
$\nabla_{\Mat{W}_s}\ell_\beta$ ; $x_i^s \in \mathbb{R}^s $, $x_j^s \in \mathbb{R}^s$  & 
$2y_k(1+r^{-1})^{-1} (\Vec{x}_i^s - \Vec{x}_j^s)({\Vec{x}^s_i}^T- {\Vec{x}^s_j}^T)\Mat{W}_s\Mat{M} 
$ \\ \hline
$\nabla_{\Mat{W}_t}\ell_\beta$ ; $x_i^s \in \mathbb{R}^s $, $x_j^t \in \mathbb{R}^t$  & 
$2y_k(1 + r^{-1})^{-1}\Vec{x}^t_j({\Vec{x}^t_j}^T\Mat{W}_t - {\Vec{x}^s_i}^T\Mat{W}_s)\Mat{M} 
$ \\ \hline
$\nabla_{\Mat{W}_t}\ell_\beta$ ; $x_i^t \in \mathbb{R}^t $, $x_j^t \in \mathbb{R}^t$  & 
$2y_k(1+r^{-1})^{-1} (\Vec{x}_i^t - \Vec{x}_j^t)({\Vec{x}^t_i}^T- {\Vec{x}^t_j}^T)\Mat{W}_t\Mat{M} 
$ \\ \hline
$\nabla_{\Mat{M}}\ell_\beta$ & $2y_k(1 + r^{-1})^{-1}(\Mat{W}_s^T{\Vec{x}^s_i}-\Mat{W}_t^T\Vec{x}^t_j)({\Vec{x}^s_i}^T\Mat{W}_s - {\Vec{x}^t_j}^T\Mat{W}_t)$ \\ \hline
$\nabla_{v_k}\ell_\beta$ & $-e^{v_k}(1 + r^{-1})^{-1}$ \\ \hline
$\nabla_{\Mat{W}_s}\mathcal{L}_{u}$  & 
$\frac{1}{p}\Mat{\Sigma}_s\Mat{W}_s\Big( 2\big(\Mat{W}_s^T\Mat{\Sigma}_s\Mat{W}_s + \Mat{W}_t^T\Mat{\Sigma}_t\Mat{W}_t \big)^{-1}  - \big(\Mat{W}_s^T\Mat{\Sigma}_s\Mat{W}_s \big)^{-1}\Big)$  \\ \hline
\end{tabular}
\end{table}

\subsection*{Derivative of Statistical loss $\mathcal{L}_u$}

The statistical loss (\ie unsupervised loss) in Eq.~\ref{eqn:generalCost} is defined using the stein divergence $\delta_s$. We have, 

\begin{align}
\small
\label{eqn:statLoss_ap}
\mathcal{L}_{u} &= \frac{1}{p} \delta_s \big( \Mat{W}_s^T \Mat{\Sigma}_s\Mat{W}_s,\Mat{W}_t^T \Mat{\Sigma}_t\Mat{W}_t \big) \nonumber\\
&= \frac{1}{p}\bigg\{ \log\det\bigg(\frac{\Mat{W}_s^T \Mat{\Sigma}_s\Mat{W}_s+\Mat{W}_t^T \Mat{\Sigma}_t\Mat{W}_t}{2}\bigg) -\frac{1}{2}\log\det(\Mat{W}_s^T \Mat{\Sigma}_s\Mat{W}_s\Mat{W}_t^T \Mat{\Sigma}_t\Mat{W}_t) \bigg\} \nonumber\\
&= \frac{1}{p}\bigg\{ \log\det\bigg(\frac{\Mat{W}_s^T \Mat{\Sigma}_s\Mat{W}_s+\Mat{W}_t^T \Mat{\Sigma}_t\Mat{W}_t}{2}\bigg) -\frac{1}{2}\log\det(\Mat{W}_s^T \Mat{\Sigma}_s\Mat{W}_s) -\frac{1}{2}\log\det(\Mat{W}_t^T \Mat{\Sigma}_t\Mat{W}_t)\bigg\},
\end{align}

\noindent where $\Mat{\Sigma}_s$ and $\Mat{\Sigma}_t$ are respectively the source and target domain covariances. 
Making use of $2.11$ of~\cite{MatrixRefManual}, we yield

\begin{align}
\nabla_{\Mat{W}_s}\log\det(\Mat{W}_s^T \Mat{\Sigma}_s\Mat{W}_s) &= 2 \Mat{\Sigma}_s\Mat{W}_s(\Mat{W}_s^T \Mat{\Sigma}_s\Mat{W}_s)^{-1}.
\end{align}

\noindent The derivative of Eq.~\ref{eqn:statLoss_ap} with respect to $\Mat{W}_s$ (or similarly for $\Mat{W}_t$\footnote{The Stein divergence is symmetric over its two arguments.}) can be obtained, 

\begin{align}
\nabla_{\Mat{W}_s}\mathcal{L}_{u} &= \frac{1}{p}\Mat{\Sigma}_s\Mat{W}_s\bigg\{\bigg(\frac{\Mat{W}_s^T \Mat{\Sigma}_s\Mat{W}_s+\Mat{W}_t^T \Mat{\Sigma}_t\Mat{W}_t}{2}\bigg)^{-1} - (\Mat{W}_s^T \Mat{\Sigma}_s\Mat{W}_s)^{-1} \bigg\} \nonumber \\
&= \frac{1}{p}\Mat{\Sigma}_s\Mat{W}_s\bigg\{2\Big(\Mat{W}_s^T \Mat{\Sigma}_s\Mat{W}_s+\Mat{W}_t^T \Mat{\Sigma}_t\Mat{W}_t\Big)^{-1} - (\Mat{W}_s^T \Mat{\Sigma}_s\Mat{W}_s)^{-1} \bigg\}
\end{align}

\noindent The derivatives are summarized in Table~\ref{tab:SummaryGrad_ap}.


\end{document}